\documentclass[runningheads]{llncs}
\usepackage{graphicx}
\usepackage[misc]{ifsym}
\usepackage{amssymb}
\usepackage{amsfonts}
\usepackage{amsmath}
\usepackage{booktabs}
\usepackage{subcaption}
\usepackage{enumitem}
\usepackage{multirow}
\usepackage[ruled, linesnumbered, vlined]{algorithm2e}
\usepackage{verbatim}
\SetKwInput{KwInput}{Input}                % Set the Input
\SetKwInput{KwOutput}{Output}
\SetKwProg{Fn}{Procedure}{}{end}
\usepackage{url}

\begin{document}
\title{Improving Few-Shot Inductive Learning on Temporal Knowledge Graphs using Confidence-Augmented Reinforcement Learning}
\toctitle{Improving Few-Shot Inductive Learning on Temporal Knowledge Graphs using Confidence-Augmented Reinforcement Learning}
\titlerunning{Few-Shot Inductive Learning on TKGs using Confidence-Augmented RL}
\author{Zifeng Ding\footnote[1]{Equal contribution.}\inst{1,2} \and Jingpei Wu\inst{\star 1} \and Zongyue 
Li\inst{1} \and \\ Yunpu Ma\inst{1,2} \and Volker Tresp(\Letter)\inst{1}}
\authorrunning{Ding et al.} % abbreviated author list (for running head)
\tocauthor{Zifeng Ding, Jingpei Wu, Zongyue Li, Yunpu Ma, Volker Tresp}
\institute{LMU Munich, Geschwister-Scholl-Platz 1, 80539 Munich, Germany
\and
Siemens AG, Otto-Hahn-Ring 6, 81739 Munich, Germany
\\
\email{zifeng.ding@campus.lmu.de, jingpei.wu@outlook.com, zongyue.li@outlook.com, cognitive.yunpu@gmail.com, Volker.Tresp@lmu.de}
}
\maketitle              % typeset the header of the contribution
\begin{abstract}
Temporal knowledge graph completion (TKGC) aims to predict the missing links among the entities in a temporal knwoledge graph (TKG). Most previous TKGC methods only consider predicting the missing links among the entities seen in the training set, while they are unable to achieve great performance in link prediction concerning newly-emerged unseen entities. Recently, a new task, i.e., TKG few-shot out-of-graph (OOG) link prediction, is proposed, where TKGC models are required to achieve great link prediction performance concerning newly-emerged entities that only have few-shot observed examples. In this work, we propose a TKGC method FITCARL that combines few-shot learning with reinforcement learning to solve this task. In FITCARL, an agent traverses through the whole TKG to search for the prediction answer. A policy network is designed to guide the search process based on the traversed path. To better address the data scarcity problem in the few-shot setting, we introduce a module that computes the confidence of each candidate action and integrate it into the policy for action selection. We also exploit the entity concept information with a novel concept regularizer to boost model performance. Experimental results show that FITCARL achieves stat-of-the-art performance on TKG few-shot OOG link prediction. Code and supplementary appendices are provided\footnote{https://github.com/ZifengDing/FITCARL/tree/main}.

\keywords{Temporal knowledge graph \and Reinforcement learning \and Few-shot learning}
\end{abstract}
\section{Introduction}
Knowledge graphs (KGs) store knowledge by representing facts in the form of triples, i.e, $(s,r,o)$, where $s$ and $o$ are the subject and object entities, and $r$ denotes the relation between them. To further specify the time validity of the facts, temporal knowledge graphs (TKGs) are introduced by using a quadruple $(s,r,o,t)$ to represent each fact, where $t$ is the valid time of this fact. In this way, TKGs are able to capture the ever-evolving knowledge over time. It has already been extensively explored to use KGs and TKGs to assist downstream tasks, e.g., question answering \cite{DBLP:conf/aaai/ZhangDKSS18,DBLP:conf/acl/SaxenaTT20,DBLP:journals/corr/abs-2208-06501} and natural language generation \cite{DBLP:conf/iclr/AmmanabroluH20,DBLP:conf/acl/LiTZWYW21}.

Since TKGs are known to be incomplete \cite{DBLP:conf/www/LeblayC18}, a large number of researches focus on proposing methods to automatically complete TKGs, i.e., temporal knowledge graph completion (TKGC). In traditional TKGC, models are given a training set consisting of a TKG containing a finite set of entities during training, and they are required to predict the missing links among the entities seen in the training set. Most previous TKGC methods, e.g., \cite{tresp2015learning,DBLP:conf/www/LeblayC18,DBLP:conf/kdd/JungJK21,ding2022simple}, achieve great success on traditional TKGC, however, they still have drawbacks. (1) Due to the ever-evolving nature of world knowledge, new unseen entities always emerge in a TKG and traditional TKGC methods fail to handle them. (2) Besides, in real-world scenarios, newly-emerged entities are usually coupled with only a few associated edges \cite{ding2022few}. Traditional TKGC methods require a large number of entity-related data examples to learn expressive entity representations, making them hard to optimally represent newly-emerged entities. To this end, recently, Ding et al. \cite{ding2022few} propose the TKG few-shot out-of-graph (OOG) link prediction (LP) task based on traditional TKGC, aiming to draw attention to studying how to achieve better LP results regarding newly-emerged TKG entities. 
% They also propose a meta-learning-based TKGC method, i.e., FILT, to solve this task by better learning the inductive representations of few-shot unseen entities. While FILT outperforms traditional TKGC methods on TKG few-shot OOG LP, it is still far from satisfactory.
% , especially in the 1-shot case.

% Reinforcement learning (RL) has gained great attention in recent years. 
% % A policy is learned to guide an RL agent to take optimal actions in order to get the maximized cumulative reward. 
% TITer \cite{sun-etal-2021-timetraveler} and CluSTeR \cite{DBLP:conf/acl/LiJGLGWC20} use RL to model the temporal path in TKGs for TKG reasoning and achieve good performance. However, they cannot handle newly-emerged few-shot entities for solving TKG few-shot OOG LP.
% They focus on TKG forecasting\footnote{TKG forecasting is different from TKGC. Explained in Section \ref{sec: rl related}.}, where they are asked to predict the missing links happening at future timestamps given the historical TKG facts. 
% Although they achieve superior performance on TKG forecasting, they are not designed for TKGC and they also cannot handle newly-emerged few-shot entities for solving TKG few-shot OOG LP.

In this work, we propose a TKGC method to improve \textbf{f}ew-shot \textbf{i}nductive learning over newly-emerged entities on \textbf{T}KGs using \textbf{c}onfidence-\textbf{a}ugmented \textbf{r}ein-forcement \textbf{l}earning (FITCARL). FITCARL is developed to solve TKG few-shot OOG LP \cite{ding2022few}. 
% Unlike TITer and CluSTeR, 
It is a meta-learning based method trained with episodic training \cite{DBLP:conf/nips/VinyalsBLKW16}. 
% and it is specifically designed to reason over few-shot unseen entities. 
For each unseen entity, FITCARL first employs a time-aware Transformer \cite{DBLP:conf/nips/VaswaniSPUJGKP17} to adaptively learn its expressive representation. Then it starts from the unseen entity and sequentially takes actions by transferring to other entities according to the observed edges associated with the current entity, following a policy parameterized by a learnable policy network. FITCARL traverses the TKG for a fixed number of steps and stops at the entity that is expected to be the LP answer. To better address the data scarcity problem in the few-shot setting, we introduce a confidence learner that computes the confidence of each candidate action and integrate it into the policy for action selection. Following \cite{ding2022few}, we also take advantage of the concept information presented in the temporal knowledge bases (TKBs) and design a novel concept regularizer. 
We summarize our contributions as follows:
(1) This is the first work using reinforcement learning-based method to reason over newly-emerged few-shot entities in TKGs and solve the TKG few-shot OOG LP task.
(2) We propose a time-aware Transformer using a time-aware positional encoding method to better utilize few-shot information in learning representations of new-emerged entities.
(3) We design a novel confidence learner to alleviate the negative impact of the data scarcity problem brought by the few-shot setting. 
(4) We propose a parameter-free concept regularizer to utilize the concept information provided by the TKBs and it demonstrates strong effectiveness.
(5) FITCARL achieves state-of-the-art performance on all datasets of TKG few-shot OOG LP and provides explainability. 

\section{Related Work}
\subsection{Knowledge Graph \& Temporal Knowledge Graph Completion}
Knowledge graph completion (KGC) methods can be summarized into two types. First type of methods focus on designing KG score functions that directly compute the plausibility scores of KG triples \cite{DBLP:conf/nips/BordesUGWY13,DBLP:conf/aaai/LinLSLZ15,DBLP:conf/nips/AbboudCLS20,DBLP:conf/icml/NickelTK11,DBLP:journals/corr/YangYHGD14a,DBLP:conf/icml/TrouillonWRGB16,DBLP:conf/emnlp/BalazevicAH19}. 
% A large number of KG score functions are translational models \cite{DBLP:conf/nips/BordesUGWY13,DBLP:conf/aaai/LinLSLZ15,DBLP:conf/nips/AbboudCLS20}. 
% They model a KG relation as a translation on a certain space from the subject entity to the object entity. 
% Another line of work is based on tensor factorization \cite{DBLP:conf/icml/NickelTK11,DBLP:journals/corr/YangYHGD14a,DBLP:conf/icml/TrouillonWRGB16,DBLP:conf/emnlp/BalazevicAH19}. 
% Given the representations of KG entities and relations, bilinear functions are used for score computation. 
Second type of KGC methods are neural-based models \cite{DBLP:conf/esws/SchlichtkrullKB18,DBLP:conf/iclr/VashishthSNT20}. Neural-based models are built by coupling KG score functions with neural structures, e.g., graph neural network (GNN). It is shown that neural structures make great contributions to enhancing the performance of KGC methods.
TKGC methods are developed by incorporating temporal reasoning techniques. A line of work aims to design time-aware KG score functions that are able to process time information \cite{DBLP:conf/www/LeblayC18,DBLP:conf/coling/XuNAYL20,DBLP:conf/aaai/SadeghianACW21,DBLP:conf/aaai/MessnerAC22,DBLP:conf/acl/ChenWLL22,DBLP:conf/cikm/ZhangZ0ZX022}. Another line of work employs neural structures to encode temporal information, where some work uses recurrent neural structures, e.g., Transformer \cite{DBLP:conf/nips/VaswaniSPUJGKP17}, to model the temporal dependencies in TKGs \cite{DBLP:conf/emnlp/WuCCH20}, and other work designs time-aware GNNs to achieve temporal reasoning by computing time-aware entity representations through aggregation \cite{DBLP:conf/kdd/JungJK21,ding2022simple}.
% \subsection{Reinforcement Learning for TKG Reasoning}
% \label{sec: rl related}
Reinforcement learning (RL) has already been used to reason TKGs, e.g., \cite{DBLP:conf/emnlp/SunZMH021,DBLP:conf/acl/LiJGLGWC20}.
TITer \cite{DBLP:conf/emnlp/SunZMH021} and CluSTeR \cite{DBLP:conf/acl/LiJGLGWC20} achieves temporal path modeling with RL. 
% An agent is employed to travel through the historical TKG facts for searching the expressive temporal information. 
However, they are traditional TKG reasoning models and are not designed to deal with few-shot unseen entities\footnote{TITer can model unseen entities, but it is not designed for few-shot setting and requires a substantial number of associated facts. Besides, both TITer and CluSTeR are TKG forecasting methods, where models are asked to predict future links given the past TKG information (different from TKGC, see Appendix B for discussion).}. 
\subsection{Inductive Learning on KGs \& TKGs} 
In recent years, inductive learning on KGs and TKGs has gained increasing interest. A series of work \cite{DBLP:conf/emnlp/XiongYCGW18,DBLP:conf/emnlp/ChenZZCC19,DBLP:conf/emnlp/ShengGCYWLX20,mirtaheri2021one,DBLP:journals/corr/abs-2205-10621} focuses on learning strong inductive representations of few-shot unseen relations using meta-learning-based approaches. These methods achieve great effectiveness, however, they are unable to deal with newly-emerged entities. Some work tries to deal with unseen entities by inductively transferring knowledge from seen to unseen entities with an auxiliary set provided during inference \cite{DBLP:conf/ijcai/HamaguchiOSM17,DBLP:conf/aaai/WangHLP19,DBLP:conf/cikm/HeWZTR20}. Their performance highly depends on the size of the auxiliary set. \cite{ding2022few} shows that with a tiny auxiliary set, these methods cannot achieve ideal performance. Besides, these methods are developed for static KGs, thus without temporal reasoning ability. On top of them, Baek et al. \cite{DBLP:conf/nips/BaekLH20} propose a more realistic task, i.e., KG few-shot OOG LP, aiming to draw attention to better studying few-shot OOG entities. They propose a model GEN that contains two GNNs
% , i.e., a transductive GNN and an inductive GNN, 
and train it with a meta-learning framework to adapt to the few-shot setting. Same as \cite{DBLP:conf/ijcai/HamaguchiOSM17,DBLP:conf/aaai/WangHLP19,DBLP:conf/cikm/HeWZTR20}, GEN does not have a temporal reasoning module, and therefore, it cannot reason TKGs. Ding et al. \cite{ding2022few} propose the TKG few-shot OOG LP task that generalizes \cite{DBLP:conf/nips/BaekLH20} to the context of TKGs. They develop a meta-learning-based model FILT that achieves temporal reasoning with a time difference-based graph encoder and mines concept-aware information from the entity concepts specified in TKBs.
Recently, another work \cite{DBLP:conf/nips/0004LSLLYA22} 
% also pays attention to few-shot inductive learning of newly-emerged entities in TKGs. 
proposes a task called few-shot TKG reasoning, aiming to ask TKG models to predict future facts for newly-emerged few-shot entities. 
% Different from TKG few-shot OOG LP, \cite{DBLP:conf/nips/0004LSLLYA22} restrains that the unseen new entities emerge over time. This means that, 
In few-shot TKG reasoning, for each newly-emerged entity, TKG models are asked to predict the unobserved associated links happening after the observed few-shot examples. Such restriction is not imposed in TKG few-shot OOG LP, meaning that TKG models should predict the unobserved links happening at any time along the time axis.
In our work, we only consider the task setting of TKG few-shot OOG LP and do not consider the setting of \cite{DBLP:conf/nips/0004LSLLYA22}.

\section{Task Formulation and Preliminaries}
\subsection{TKG Few-Shot Out-of-Graph Link Prediction}
% \begin{definition}
\textbf{Definition 1 (TKG Few-Shot OOG LP).} Assume we have a background TKG $\mathcal{G}_{\text{back}} = \{(s,r,o,t)|s,o \in \mathcal{E}_{\text{back}}, r \in \mathcal{R}, t \in \mathcal{T}\} \subseteq \mathcal{E}_{\text{back}} \times \mathcal{R} \times \mathcal{E}_{\text{back}} \times \mathcal{T}$, where $\mathcal{E}_{\text{back}}$, $\mathcal{R}$, $\mathcal{T}$ denote a finite set of seen entities, relations and timestamps, respectively. An unseen entity $e'$ is an entity $e' \in \mathcal{E'}$ and $\mathcal{E}' \cap \mathcal{E}_{\text{back}} = \emptyset$. 
For each $e' \in \mathcal{E}'$, given $K$ observed $e'$ associated TKG facts $(e', r, \Tilde{e}, t)$ (or $(\Tilde{e}, r, e', t)$), 
where $\Tilde{e} \in (\mathcal{E}_{\text{back}} \cup \mathcal{E}')$, $r \in \mathcal{R}$, $t \in \mathcal{T}$, 
TKG few-shot OOG LP asks models to predict the missing entities of LP queries $(e', r_q,?, t_q)$ (or $(?, r_q, e', t_q)$) derived from unobserved TKG facts containing $e'$ ($r_q \in \mathcal{R}$, $t_q \in \mathcal{T}$). $K$ is a small number denoting shot size, e.g., 1 or 3. 

Ding et al. \cite{ding2022few} formulate TKG few-shot OOG LP into a meta-learning problem and 
uses episodic training \cite{DBLP:conf/nips/VinyalsBLKW16} to train its model. 
For a TKG $\mathcal{G}\subseteq \mathcal{E} \times \mathcal{R} \times \mathcal{E} \times \mathcal{T}$, 
they split its entities into background (seen) entities $\mathcal{E}_{\text{back}}$ 
and unseen entities $\mathcal{E}'$, where $\mathcal{E}' \cap \mathcal{E}_{\text{back}} = \emptyset$ 
and $\mathcal{E} = (\mathcal{E}_{\text{back}} \cup \mathcal{E}')$. A background TKG 
$\mathcal{G}_{\text{back}} \subseteq \mathcal{E}_{\text{back}} \times \mathcal{R} \times \mathcal{E}_{\text{back}} \times \mathcal{T}$ 
is constructed by including all the TKG facts that do not contain unseen entities. 
Then, unseen entities $\mathcal{E}'$ are further split into three non-overlapped 
groups $\mathcal{E}'_\text{meta-train}$, $\mathcal{E}'_\text{meta-valid}$ and $\mathcal{E}'_\text{meta-test}$. 
The union of all the facts associated to each group's entities forms the corresponding 
meta-learning set, e.g., 
the meta-training set $\mathbb{T}_{\text{meta-train}}$ is formulated as 
$\{(e',r, \Tilde{e},t)|\Tilde{e} \in \mathcal{E}, r \in \mathcal{R}, e' \in \mathcal{E}'_{\text{meta-train}}, t \in \mathcal{T}\} \cup \{(\Tilde{e},r, e',t)|\Tilde{e} \in \mathcal{E}, r \in \mathcal{R}, e' \in \mathcal{E}'_{\text{meta-train}}, t \in \mathcal{T}\}$. 
Ding et al. ensure that there exists no link between every two of the meta-learning sets. 
During meta-training, models are trained over a number of episodes, where a training task $T$ is sampled in each episode. 
For each task $T$, $N$ unseen entities $\mathcal{E}_T$ are sampled 
from $\mathcal{E}'_\text{meta-train}$. For each $e' \in \mathcal{E}_T$, $K$ associated 
facts are sampled to form a support set $Sup_{e'} = \{(e',r_i,\Tilde{e}_i,t_i) \ \text{or}\  (\Tilde{e}_i,r_i,e',t_i)|\Tilde{e}_i \in (\mathcal{E}_{\text{back}} \cup \mathcal{E}'), r_i \in \mathcal{R}, t_i \in \mathcal{T}\}^K_{i=1}$, 
and the rest of its associated facts are taken as its query set 
$Que_{e'} = \{(e',r_i,\Tilde{e}_i,t_i) \ \text{or}\  (\Tilde{e}_i,r_i,e',t_i)|\Tilde{e}_i \in (\mathcal{E}_{\text{back}} \cup \mathcal{E}'), r_i \in \mathcal{R}, t_i \in \mathcal{T}\}^{M_{e'}}_{i=K+1}$, 
where $M_{e'}$ denotes the number of $e'$'s associated facts. Models are asked to 
simultaneously perform LP over $Que_{e'}$ for each $e' \in \mathcal{E}_T$, 
given their $Sup_{e'}$ and $\mathcal{G}_{\text{back}}$. After meta-training, models are 
validated with a meta-validation set $\mathbb{T}_{\text{meta-valid}}$ and tested with 
a meta-test set $\mathbb{T}_{\text{meta-test}}$. 
In our work, we also train FITCARL in the same way as \cite{ding2022few} with episodic training on the same meta-learning problem. 
\begin{figure}[htbp]
    \centering
    \includegraphics[width=0.95\textwidth]{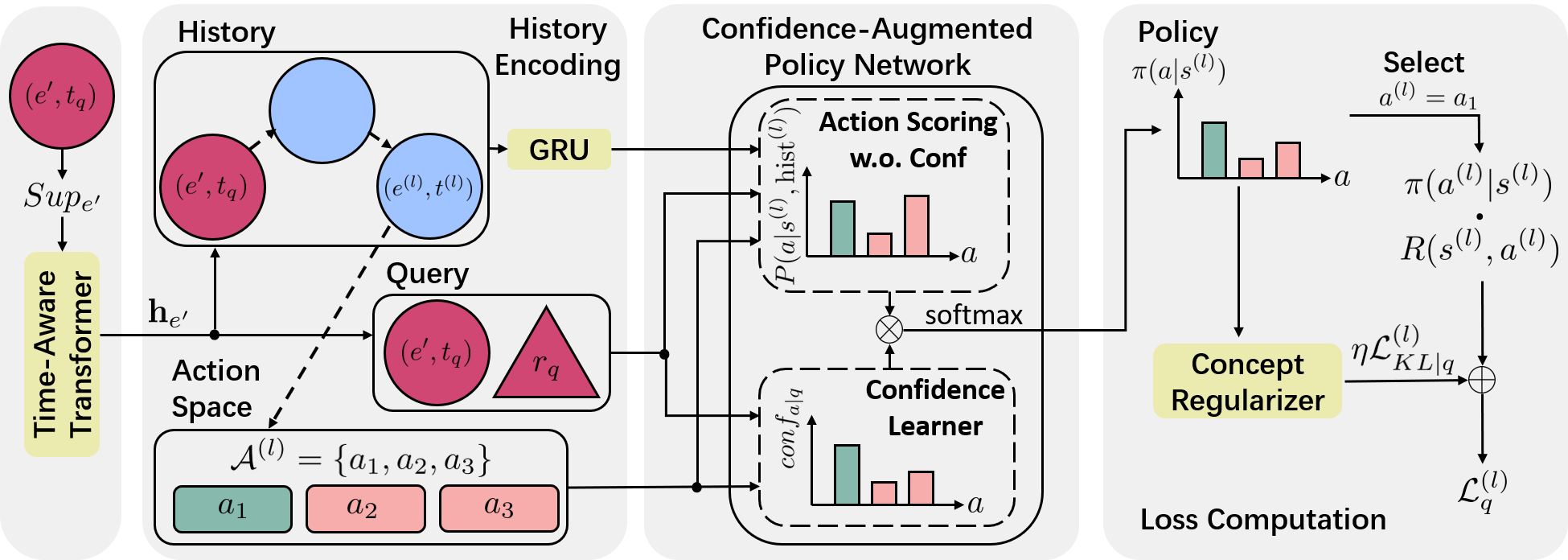}
    \caption{Overview of FITCARL. To do prediction over the LP query $q = (e', r_q, ?, t_q)$, FITCARL first learns $\mathbf{h}_{e'}$ from a time-aware Transformer. It is then used in history encoding (with GRU) and policy network. To search for the answer, FITCARL starts from node $(e', t_q)$. It goes to $(e^{(l)}, t^{(l)})$, state $s^{(l)}$, at step $l$. 
    % According to search history and $q$, 
    It computes a policy using a confidence-augmented policy network. Assume FITCARL selects action $a_1$ in current action space $\mathcal{A}^{(l)}$ as the current action $a^{(l)}$. 
    % $a_1$'s probability in policy is first coupled with $a_1$'s reward $R(s^{(l)}, a^{(l)})$ to compute a loss at step $l$. 
    We compute a loss $\mathcal{L}_q^{(l)}$ at step $l$, considering $a_1$'s probability in policy and reward $R(s^{(l)}, a^{(l)})$, as well as an extra regularization loss $\eta \mathcal{L}_{KL|q}^{(l)}$ computed by a concept regularizer.
    % A concept regularizer computes an extra loss $\eta \mathcal{L}_{KL|q}^{(l)}$ for regularization. 
    % Total loss at step $l$ is denoted as $\mathcal{L}_q^{(l)}$.
    Please refer to Section \ref{sec: unseen entity learning}, \ref{sec: RL framework} and \ref{sec: policy} for details.}
    \label{fig: model}
\end{figure}
\subsection{Concepts for Temporal Knowledge Graph Entities}
\cite{ding2022few} extracts the concepts of TKG entities by exploring the associated TKBs. 
Entity concepts describe the characteristics of entities. For example, 
in the Integrated Crisis Early Warning System (ICEWS) database \cite{DVN/28075_2015}, 
the entity \textit{Air Force (Canada)} is described with the following concepts: 
\textit{Air Force}, \textit{Military} and \textit{Government}. Ding et al. propose three 
ICEWS-based datasets for TKG few-shot OOG LP and manage to couple every entity 
with its unique concepts. 
% It is shown that concept-aware information can be used to aid the TKG model to better 
% learn newly-emerged entities. 
We use $\mathcal{C}$ to denote all the concepts existing in a TKG and 
$\mathcal{C}_e$ to denote $e$'s concepts.
\section{The Proposed FITCARL Model}
Given the support set $Sup_{e'} = \{(e',r_i,\Tilde{e}_i,t_i) \ \text{or}\  (\Tilde{e}_i,r_i,e',t_i)\}^K_{i=1}$ of $e' \in \mathcal{E}'$, 
assume we want to predict the missing entity from the LP query 
$q = (e', r_q, ?, t_q)$ derived from a query quadruple\footnote{For each query quadruple in the form of $(\Tilde{e}_q, r_q, e', t_q)$, we derive its LP query as $(e', r_q^{-1}, ?, t_q)$. $r_q^{-1}$ is $r_q$'s inverse relation. The agent always starts from $(e', t_q)$.} $(e', r_q, \Tilde{e}_q, t_q) \in Que_{e'}$. 
To achieve this, FITCARL first learns a representation $\mathbf{h}_{e'} \in \mathbb{R}^d $ 
($d$ is dimension size) for $e'$ (Section \ref{sec: unseen entity learning}). 
Then it employs an RL agent that starts from the node $(e', t_q)$ and sequentially 
takes actions by traversing to other nodes (in the form of (\textit{entity}, \textit{timestamp})) following a policy (Section \ref{sec: RL framework} and \ref{sec: policy}). 
After $L$ traverse steps, the agent is expected to stop at a target node 
containing $\Tilde{e}_q$. 
% The agent can traverse from one node to another only when there exists a temporal edge between them. 
Fig. \ref{fig: model} shows an overview of FITCARL during training, showing how it computes loss $\mathcal{L}_q^{(l)}$ at step $l$.
\subsection{Learning Unseen Entities with Time-Aware Transformer}
\label{sec: unseen entity learning}
We follow FILT \cite{ding2022few} and use the entity and relation representations pre-trained with ComplEx 
\cite{DBLP:conf/icml/TrouillonWRGB16} for model initialization. 
Note that pre-training only considers all the background TKG facts, i.e., $\mathcal{G}_{\text{back}}$. 
To learn $\mathbf{h}_{e'}$, we start from learning $K$ separate 
meta-representations. Given $Sup_{e'}$, we transform every support quadruple whose form is
$(e',r_i,\Tilde{e}_i,t_i)$ to $(\Tilde{e}_i,r_i^{-1},e',t_i)$, where $r_i^{-1}$ denotes the 
inverse relation\footnote{Both original and inverse relations are trained in pre-training.}
% We directly use the pre-trained representations of both types of relations trained in FILT. 
% The inverse of an inverse relation $r^{-1}$ corresponds to the original relation $r$.} 
of $r_i$. 
Then we create a temporal neighborhood 
$\mathcal{N}_{e'} = \{(\Tilde{e}_i, r_i, t_i)| (\Tilde{e}_i, r_i, e', t_i) \in Sup_{e'} \  \text{or} \  (e', r_i^{-1}, \Tilde{e}_i, t_i)\in Sup_{e'}\}$ for $e'$ based on $Sup_{e'}$, where $|\mathcal{N}_{e'}| = K$. We compute a meta-representation $\mathbf{h}_{e'}^{i}$ from each temporal neighbor $(\Tilde{e}_i, r_i, t_i)$ as $\mathbf{h}_{e'}^{i} = f(\mathbf{h}_{\Tilde{e}_i}\| \mathbf{h}_{r_i})$, where
% \begin{equation}
%     \mathbf{h}_{e'}^{i} = f(\mathbf{h}_{\Tilde{e}_i}\| \mathbf{h}_{r_i}).
% \end{equation}
$\mathbf{h}_{r_i} \in \mathbb{R}^d$ is the representation of the relation $r_i$ and $\|$ is the concatenation operation. 
% \subsubsection{Time-Aware Transformer}
% \label{sec: time aware transformer}

We collect $\{\mathbf{h}_{e'}^{i}\}_{i=1}^K$ and use a time-aware Transformer to compute a contextualized representation $\mathbf{h}_{e'}$. We treat each temporal neighbor $(\Tilde{e}_i, r_i, t_i) \in \mathcal{N}_{e'}$ as a token and the corresponding meta-representation $\mathbf{h}_{e'}^{i}$ as its token representation. We concatenate the classification ([CLS]) token with the temporal neighbors in $\mathcal{N}_{e'}$ as a sequence and input it into a Transformer, where the sequence length is $K+1$. The order of temporal neighbors is decided by the sampling order of support quadruples.
% In LMs, each token has its own position representation that is used to indicate its position in the sequence. Natural language texts contain words that locate in an order following syntactic structures. Therefore, the learned position representations of LMs can be shared across different input texts. 

To better utilize temporal information from temporal neighbors, we propose a time-aware positional encoding method. For any two tokens $u,v$ in the input sequence, we compute the time difference $t_u - t_v$ between their associated timestamps,
% of the $u^\text{th}$ and $v^\text{th}$ tokens in the input sequence
and then map it into a time-difference representation $\mathbf{h}_{t_u-t_v} \in \mathbb{R}^d$,
\begin{equation}
\label{eq: time encoding}
    \mathbf{h}_{t_u-t_v} = \sqrt{\frac{1}{d}}[cos(\omega_1(t_u-t_v)+\phi_1), ..., cos(\omega_{d}(t_u-t_v)+ \phi_{d}))].
\end{equation} 
$\omega_1$ to $\omega_{d}$ and $\phi_1$ to $\phi_{d}$ are trainable parameters. 
% We derive the time-aware position representation for each token (temporal neighbor) as follows.
% Inspired by T5 \cite{DBLP:journals/jmlr/RaffelSRLNMZLL20}, 
The timestamp for each temporal neighbor is $t_i$ and we set the timestamp of the [CLS] token to the
query timestamp $t_q$ since we would like to use the learned $\mathbf{h}_{e'}$ to predict the LP query happening at $t_q$.
The attention $\text{att}_{u,v}$ of any token $v$ to token $u$ in an attention layer of our 
time-aware Transformer is written as
\begin{equation}
\label{eq: time att transformer}
\begin{aligned}
    &\text{att}_{u,v} = \frac{\text{exp}(\alpha_{u,v})}{\sum_{k=1}^{K+1} \text{exp}(\alpha_{u,k})},\\
    &\alpha_{u,v} = \frac{1}{\sqrt{d}}( \mathbf{W}_{TrQ} \mathbf{h}_u)^\top(\mathbf{W}_{TrK} \mathbf{h}_v) + {\mathbf{w}_{Pos}}^\top \mathbf{h}_{t_u - t_v}.
\end{aligned}    
\end{equation}
\begin{figure}[htbp]
    \centering
    \includegraphics[width=0.9\textwidth]{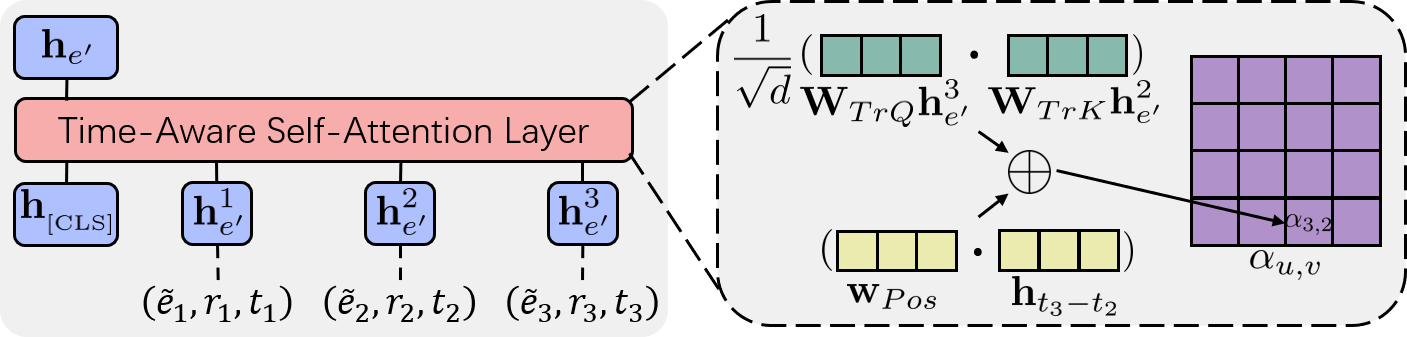}
    \caption{Time-aware Transformer with one attention layer for learning unseen entity representation in the 3-shot case.}
    \label{fig: ta transformer}
\end{figure}
$\mathbf{h}_u, \mathbf{h}_v \in \mathbb{R}^d$ are the input representations of token $u,v$ into this
attention layer. $\mathbf{W}_{TrQ}, \mathbf{W}_{TrK} \in \mathbb{R}^{d \times d}$ are the weight matrices 
following original definition in \cite{DBLP:conf/nips/VaswaniSPUJGKP17}. 
$\mathbf{w}_{Pos} \in \mathbb{R}^d$ is a parameter that maps $\mathbf{h}_{t_u - t_v}$ to a scalar 
representing time-aware relative position from token $v$ to $u$. We use several attention layers and 
also employ multi-head attention to increase model expressiveness. 
% In previous work that uses Transformer to encode natural language texts, e.g., large language models (LMs) like BERT \cite{DBLP:conf/naacl/DevlinCLT19}, the Transformer output of the classification ([CLS]) token is taken as the representation of the whole input sequence, e.g., a natural language sentence. 
The output representation of the [CLS] token from the last attention layer is taken as $\mathbf{h}_{e'}$. Fig. \ref{fig: ta transformer} illustrates how the time-aware Transformer learns $\mathbf{h}_{e'}$ in the 3-shot case.
% In our work, the temporal neighbors come from the sampled support set, where the order of the sampled support quadruples is always random and carries no information. To address this problem, we design a time-aware positional encoding method.
% The order of the input sequence is decided when the support quadruples are sampled. In meta-validation and meta-test, it corres
% where the length of it is $K$. Figure \ref{} illustrates how we learn $\mathbf{h}_{e'}$ from $\{\mathbf{h}_{e'}^{i}\}_{i=1}^K$.
\subsection{Reinforcement Learning Framework}
\label{sec: RL framework}
% Assume we want to perform LP on an LP query $q = (e', r_q, ?, t_q)$ (derived from a query quadruple $(e', r_q, \Tilde{e}_q, t_q) \in Que_{e'}$). Let $(e', t_q)$ denotes the node representing $e'$ at $t_q$. Similar to \cite{sun-etal-2021-timetraveler}, FITCARL employs an agent that starts from $(e', t_q)$ and sequentially takes actions by traversing to other nodes (in the form of (\textit{entity}, \textit{timestamp})), following a policy. After a number of traverse steps, the agent is expected to stop at a target node corresponding to the ground truth missing entity $\Tilde{e}_q$. The agent can traverse from one node to another only when there exists a temporal edge between them. Figure \ref{} shows an example of how FITCARL seaerches for the prediction answer. 
We formulate the RL process as a Markov Decision Process and we introduce its elements as follows.
\textbf{(1) States:} Let $\mathcal{S}$ be a state space. A state is denoted as 
$s^{(l)} = (e^{(l)}, t^{(l)}, e', r_q, t_q) \in \mathcal{S}$. $(e^{(l)}, t^{(l)})$ is the node that is 
visited by the agent at step $l$ and $e', r_q, t_q$ are taken from the LP query $(e', r_q, ?, t_q)$. 
The agent starts from $(e', t_q)$, and thus $s^{(0)} = (e', t_q, e', r_q, t_q)$.
% \paragraph{Actions.} Let $\mathcal{A}$ denote an action space and $\mathcal{A}^{(l)} \subset \mathcal{A}$ denotes the action space at step $l$. $\mathcal{A}^{(l)} = \{(r, e, t)| (e^{(l)}, r, e, t) \in (\mathcal{G}_{\text{back}} \cup \bigcup_{e'' \in \mathcal{E}_T} Sup_{e''}), r \in \mathcal{R}, e \in (\mathcal{E}_{\text{back}} \cup \mathcal{E}_T), t \in \mathcal{T}\}$ contains all the possible outgoing edges starting from the node $(e^{(l)}, t^{(l)})$ at current step. 
\textbf{(2) Actions:} Let $\mathcal{A}$ denote an action space and $\mathcal{A}^{(l)} \subset \mathcal{A}$ denotes the action space at step $l$. $\mathcal{A}^{(l)}$ is sampled from all the possible outgoing edges starting from $(e^{(l)}, t^{(l)})$, i.e., $\{a=(r, e, t)| (e^{(l)}, r, e, t) \in (\mathcal{G}_{\text{back}} \cup \bigcup_{e'' \in \mathcal{E}_T} Sup_{e''}), r \in \mathcal{R}, e \in (\mathcal{E}_{\text{back}} \cup \mathcal{E}_T), t \in \mathcal{T}\}$. 
We do sampling because if $e^{(l)} \in \mathcal{E}_{\text{back}}$, there probably exist lots of outgoing edges 
in $\mathcal{G}_{\text{back}}$. If we include all of them into $\mathcal{A}^{(l)}$, they will lead to an 
excessive consumption of memory and cause out-of-memory problem on hardware devices. 
We sample $\mathcal{A}^{(l)}$ in a time-adaptive manner. For each outgoing edge $(r,e,t)$, we compute a score 
${\mathbf{w}_{\Delta t}}^\top \mathbf{h}_{t_q - t}$, where $\mathbf{w}_{\Delta t} \in \mathbb{R}^d$ is a 
time modeling weight and $\mathbf{h}_{t_q - t}$ is the representation denoting the time difference 
$t_q-t$. $\mathbf{h}_{t_q - t}$ is computed as in Equation \ref{eq: time encoding} with shared parameters. 
We rank the scores of outgoing edges in the descending order and 
take a fixed number of top-ranked edges as $\mathcal{A}^{(l)}$. We also include one self-loop action in each $\mathcal{A}^{(l)}$ that makes the agent stay at the current node.
% In this way, 
% FITCARL adaptively chooses the most temporally contributive edges as the candidate actions.
% Since each entity in $\mathcal{E}_{\text{back}}$ has a large number of observed facts in $\mathcal{G}$, there probably exist lots of outgoing edges $(e^{(l)}, t^{(l)})$. If we include all of them into $\mathcal{A}^{(l)}$, it will lead to an excessive consumption of memory and cause out-of-memory problem on hardware devices. from We sample $\mathcal{A}^{(l)}$ according in a time-adaptive manner. We first compute 
% Note that we do not impose constraint to the timestamps of the actions in $\mathcal{A}^{(l)}$, leading to a large size of $\mathcal{A}^{(l)}$. Previous work \cite{sun-etal-2021-timetraveler,DBLP:conf/acl/LiJGLGWC20} avoids large action spaces by only keeping a fixed number of temporally-nearest actions corresponding to the TKG facts happening prior to $t^{(l)}$. However, in FITCARL, we keep all the actions corresponding to the TKG facts happening at any timestamp and use a time-adaptive guidance module to let model adaptively choose the most contributive actions. This helps FITCARL better utilize temporal information and achieve greater performance (see Section \ref{} and \ref{} for detailed discussions).
\textbf{(3) Transition:} A transition fuction $\delta$
% $\delta: \mathcal{S} \times \mathcal{A} \rightarrow \mathcal{S}$ 
is used to transfer from one state to another, i.e., $\delta(s^{(l)}, a^{(l)}) = s^{(l+1)} = (e^{(l+1)}, t^{(l+1)}, e', r_q, t_q)$, according to the selected action $a^{(l)}$.
\textbf{(4) Rewards:} 
% Unlike previous work \cite{sun-etal-2021-timetraveler,DBLP:conf/acl/LiJGLGWC20} that only gives the agent a terminal reward if it stops at the ground truth missing entity $\Tilde{e}_q$ at the end of search, 
% in FITCARL, 
We give the agent a reward at each step of state transition and consider a cumulative reward for the whole 
searching process. The reward of doing a candidate action $a \in \mathcal{A}^{(l)}$ at step $l$ is given as 
$R(s^{(l)}, a) = \text{Sigmoid}\left(\theta - \left\|\mathbf{h}_{\Tilde{e}_q} - \mathbf{h}_{e_{a}}\right\|_2\right).$
% \begin{equation}
%     R(s^{(l)}, a) = \text{Sigmoid}\left(\theta - \left\|\mathbf{h}_{\Tilde{e}_q} - \mathbf{h}_{e_{a}}\right\|_2\right).
% \end{equation}
$\theta$ is a hyperparameter adjusting the range of reward. $\mathbf{h}_{e_{a}}$ denotes the representation of entity $e_{a}$ 
selected in the action $a=(r_a, e_a, t_a)$. $\|\cdot\|_2$ is the L2 norm. 
The closer $e_{a}$ is to $\Tilde{e}_q$, the greater reward the agent gets if it does action $a$.
% The motivation of this change is explained as follows. For each newly-emerged entity, the number of observed associated edges is extremely small, e.g., 1 or 3. In this case, it is much harder for the agent to transfer to $\Tilde{e}_q$, since it is not provided with an extensive search space (action space) at the start of the searching process. If the observed few-shot examples provide little valuable information for predicting $\Tilde{e}_q$, the agent may traverse in the wrong direction.
\subsection{Confidence-Augmented Policy Network}
\label{sec: policy}
We design a confidence-augmented policy network that calculates the probability distribution over all the 
candidate actions $\mathcal{A}^{(l)}$ at the search step $l$, according to the current state $s^{(l)}$, 
the search history $\text{hist}^{(l)} = ((e', t_q), r^{(1)}, (e^{(1)}, t^{(1)}), ..., r^{(l)}, \\(e^{(l)}, t^{(l)}))$, 
and the confidence $\text{conf}_{a|q}$ of each $a \in \mathcal{A}^{(l)}$. During the search, we represent each visited node with a time-aware representation related to the LP query $q$. For example, for the node $(e^{(l)}, t^{(l)})$ visited at step $l$, we compute its representation as $\mathbf{h}_{(e^{(l)}, t^{(l)})} = \mathbf{h}_{e^{(l)}} \| \mathbf{h}_{t_q - t^{(l)}}$.
% \begin{equation}
%     \mathbf{h}_{(e^{(l)}, t^{(l)})} = \mathbf{h}_{e^{(l)}} \| \mathbf{h}_{t_q - t^{(l)}}.
% \end{equation}
$\mathbf{h}_{t_q - t^{(l)}}$ is computed as same in Equation \ref{eq: time encoding} and parameters are shared.
\subsubsection{Encoding Search History}
% We learn the representation of the search history $\text{hist}^{(l)}$ as
% \begin{equation}
%         \mathbf{h}_{\text{hist}^{(l)}} = \text{GRU}\left( \left(\mathbf{h}_{{r}^{(l)}} \| \mathbf{h}_{(e^{(l)}, t^{(l)})}\right),\mathbf{h}_{\text{hist}^{(l-1)}}\right),\ \text{where} \mathbf{h}_{\text{hist}^{(0)}} = \text{GRU}\left( \left(\mathbf{h}_{{r}_{\text{dummy}}} \| \mathbf{h}_{(e', t_q)}\right), \mathbf{0}\right).
% \end{equation}
The search history $\text{hist}^{(l)}$ is encoded as
\begin{equation}
\begin{aligned}
        &\mathbf{h}_{\text{hist}^{(l)}} = \text{GRU}\left( \left(\mathbf{h}_{{r}^{(l)}} \| \mathbf{h}_{(e^{(l)}, t^{(l)})}\right),\mathbf{h}_{\text{hist}^{(l-1)}}\right),\\
        &\mathbf{h}_{\text{hist}^{(0)}} = \text{GRU}\left( \left(\mathbf{h}_{{r}_{\text{dummy}}} \| \mathbf{h}_{(e', t_q)}\right), \mathbf{0}\right).
\end{aligned}
\end{equation}
GRU is a gated recurrent unit \cite{DBLP:conf/emnlp/ChoMGBBSB14}. 
$\mathbf{h}_{\text{hist}^{(0)}} \in \mathbb{R}^{3d}$ is the initial hidden state of GRU 
and $\mathbf{h}_{{r}_{\text{dummy}}}\in \mathbb{R}^{d}$ is the representation of a dummy relation 
for GRU initialization. $\mathbf{h}_{(e', t_q)}$ is the time-aware representation of the starting node $(e', t_q)$. 
% derived from the LP query $q$.
\subsubsection{Confidence-Aware Action Scoring}
We design a score function for computing the probability of selecting each candidate action $a \in \mathcal{A}^{(l)}$. Assume $a=(r_a, e_a, t_a)$, where $(e^{(l)}, r_a, e_a, t_a) \in (\mathcal{G}_{\text{back}} \cup \bigcup_{e'' \in \mathcal{E}_T} Sup_{e''})$. We first compute an attentional feature $\mathbf{h}_{\text{hist}^{(l)}, q|a}$ that extracts the information highly-related to action $a$ from the visited search history $\text{hist}^{(l)}$ and the LP query $q$.
\begin{equation}
\label{eq: attentional feature}
\begin{aligned}
        &\mathbf{h}_{\text{hist}^{(l)}, q|a} = \text{att}_{\text{hist}^{(l)}, a} \cdot \Bar{\mathbf{h}}_{\text{hist}^{(l)}} + \text{att}_{q, a} \cdot \Bar{\mathbf{h}}_{q}, \\
        &\Bar{\mathbf{h}}_{\text{hist}^{(l)}} = {\mathbf{W}_1}^\top\mathbf{h}_{\text{hist}^{(l)}}, \quad
        \Bar{\mathbf{h}}_{q} = {\mathbf{W}_2}^\top\left(\mathbf{h}_{r_q}\|\mathbf{h}_{(e', t_q)}\right).
\end{aligned}
\end{equation}
$\mathbf{W}_1, \mathbf{W}_2 \in \mathbb{R}^{2d \times 3d}$ are two weight matrices. $\mathbf{h}_{r_q}$ is the representation of the query relation $r_q$. $\text{att}_{\text{hist}^{(l)},a}$ and $\text{att}_{q,a}$ are two attentional weights that are defined as
% distinguishing the importance of action $a$ to the search history and the query $q$, respectively. Formally, they are defined as
\begin{equation}
    \text{att}_{\text{hist}^{(l)},a} = \frac{\text{exp}(\phi_{\text{hist}^{(l)},a})} {\text{exp}(\phi_{\text{hist}^{(l)},a})+\text{exp}(\phi_{q,a})}, \text{att}_{q,a} = \frac{\text{exp}(\phi_{q,a})}{\text{exp}(\phi_{\text{hist}^{(l)},a})+\text{exp}(\phi_{q,a})},
\end{equation}
where
\begin{equation}
    \begin{aligned}
        \phi_{\text{hist}^{(l)},a} = {\Bar{\mathbf{h}}_a}^\top \Bar{\mathbf{h}}_{\text{hist}^{(l)}} &+ \mathbf{w}_{\Delta t}^\top \mathbf{h}_{t_a - t^{(l)}}, \ 
        \phi_{q,a} = {\Bar{\mathbf{h}}_a}^\top \Bar{\mathbf{h}}_{q} + \mathbf{w}_{\Delta t}^\top \mathbf{h}_{t_a - t_q},\\
        &\Bar{\mathbf{h}}_a = {\mathbf{W}_3}^\top\left(\mathbf{h}_{r_a}\|\mathbf{h}_{(e_a, t_a)}\right).
    \end{aligned}
\end{equation}
$\mathbf{W}_3 \in \mathbb{R}^{2d\times3d}$ is a weight matrix. $\mathbf{h}_{r_a}$ is the representation of $r_a$. $\mathbf{h}_{(e_a, t_a)}$ is the time-aware representation of node $(e_a, t_a)$ from action $a$. 
$\mathbf{w}_{\Delta t}$ maps time differences to a scalar indicating how 
temporally important is the action $a$ to the history and the query $q$. 
We take $t^{(l)}$ as search history's timestamp because it is the timestamp of the node where the search stops.
%  the current step $l$.
% We also enable our model to directly consider the temporal information $\mathbf{w}_{\Delta t}$
Before considering confidence, we compute a probability for each candidate action $a \in \mathcal{A}^{(l)}$ at step $l$
\begin{equation}
\label{eq: prob before confidence}
    P(a|s^{(l)}, \text{hist}^{(l)}) = \frac{\text{exp}({\Bar{\mathbf{h}}_a}^\top \mathbf{W}_4 \mathbf{h}_{\text{hist}^{(l)}, q|a})}{\sum_{a' \in \mathcal{A}^{(l)}} \text{exp}({\Bar{\mathbf{h}}_{a'}}^\top \mathbf{W}_4 \mathbf{h}_{\text{hist}^{(l)}, q|a'})},
\end{equation}
where $\mathbf{W}_4 \in \mathbb{R}^{2d \times 2d}$ is a weight matrix. The probability of each action $a$ is 
decided by its associated node $(e_a, t_a)$ and the attentional feature $\mathbf{h}_{\text{hist}^{(l)}, q|a}$ 
that adaptively selects the information highly-related to $a$.

In TKG few-shot OOG LP, only a small number of $K$ edges associated to each unseen entity are observed. 
This leads to an incomprehensive action space $\mathcal{A}^{(0)}$ at the start of search because our agent 
starts travelling from node $(e', t_q)$ and $|\mathcal{A}^{(0)}|=K$ is extremely tiny. 
Besides, since there exist plenty of unseen entities in $\mathcal{E}_T$, it is highly probable that the agent 
travels to the nodes with other unseen entities during the search, causing it sequentially experience multiple tiny action spaces. As the number of the experienced incomprehensive action spaces increases, more noise will be introduced in history encoding.
From Equation \ref{eq: attentional feature} to \ref{eq: prob before confidence}, 
we show that we heavily rely on the search history for computing candidate action probabilities.
% incomprehensive action spaces would introduce increasing noise. 
% in computing action probability.
% which leads to a much worse performance. 
To address this problem, we design a confidence learner that learns the confidence $\text{conf}_{a|q}$ of each
$a \in \mathcal{A}^{(l)}$, independent of the search history. The form of confidence learner is inspired by a KG score 
function TuckER \cite{DBLP:conf/emnlp/BalazevicAH19}.
\begin{equation}
\label{eq: confidence}
\begin{aligned}
        &\text{conf}_{a|q} = \frac{\text{exp}(\psi_{a|q})}{\sum_{a' \in \mathcal{A}^{(l)}} \text{exp}(\psi_{a'|q})}, \ \text{where}\ \psi_{a|q} = \mathcal{W} \times_1 \mathbf{h}_{(e', t_q)} \times_2 \mathbf{h}_{r_q} \times_3 \mathbf{h}_{(e_a, t_a)}.
\end{aligned}
\end{equation}
$\mathcal{W} \in \mathbb{R}^{2d \times d \times 2d}$ is a learnable core tensor introduced in \cite{DBLP:conf/emnlp/BalazevicAH19}. As defined in tucker decomposition \cite{tucker64extension}, $\times_1, \times_2, \times_3$ are three operators indicating the tensor product in three different modes (see \cite{DBLP:conf/emnlp/BalazevicAH19,tucker64extension} for detailed explanations). Equation \ref{eq: confidence} can be interpreted as another action scoring process that is irrelevant to the search history.
% first computing the score of the TKG quadruple $(e', r_q, e_a, t_q)$ for each action $a$, and then perform softmax, 
% without considering the search history. 
% TuckER is originally We use time-aware representations of node $(e', t_q)$ and $(e_q, t_q)$ to achieve temporal reasoning
% A KG scoring function is learned to return high scores for ground truth quadruples and low scores for false ones. We aim to
% If the score of $(e', r_q, e_a, t_q)$ is high, then it implies that choosing action $a$ is sensible 
If $\psi_{a|q}$ is high, then it implies that choosing action $a$ is sensible 
and $e_a$ is likely to resemble the ground truth missing entity $\Tilde{e}_q$. 
Accordingly, the candidate action $a$ will be assigned a great confidence. In this way, 
we alleviate the negative influence of cascaded noise introduced by multiple tiny action spaces in the search history.
% and guide the agent to search along the node path 
% that is most likely to find the ground truth missing entity $\Tilde{e}_q$. 
The policy $\pi(a|s^{(l)})$ at step $l$ is defined as
\begin{equation}
\label{eq: policy}
    \pi(a|s^{(l)}) = \frac{\text{exp}(P(a|s^{(l)}, \text{hist}^{(l)}) \cdot \text{conf}_{a|q})}{\sum_{a' \in \mathcal{A}^{(l)}} \text{exp}(P(a'|s^{(l)}, \text{hist}^{(l)}) \cdot \text{conf}_{a'|q})}
\end{equation}
\subsection{Concept Regularizer}
In the background TKG $\mathcal{G}_{\text{back}}$, the object entities of each relation 
conform to a unique distribution. For each relation $r \in \mathcal{R}$, we track all the 
TKG facts containing $r$ in $\mathcal{G}_{\text{back}}$, and pick out all their object entities 
$\mathcal{E}_r$ ($\mathcal{E}_r \in \mathcal{E}_{\text{back}}$) together with 
their concepts $\{\mathcal{C}_e|e \in \mathcal{E}_r\}$. We sum up the number of appearance $n_c$ of each concept $c$ and compute a probability $P(c|r)$ denoting how probable it is to see $c$ when we perform object prediction\footnote{All LP queries are transformed into object prediction in TKG few-shot OOG LP.} over the LP queries concerning $r$. For example, for $r$, $\mathcal{E}_r = \{e_1, e_2\}$ and $\mathcal{C}_{e_1} = \{c_1, c_2\}$, $\mathcal{C}_{e_2} = \{c_2\}$. The probability $P(c_1|r) = n_{c_1}/\sum_{c\in \mathcal{C}}n_{c} = 1/3$, $P(c_2|r) = n_{e_2}/\sum_{c\in \mathcal{C}}n_{c} = 2/3$. 
% \begin{figure}
%     \centering
%     \includegraphics[width=0.9\textwidth]{pics/Concept_Regularizer.png}
%     \caption{Concept regularizer. $P(a_1|\mathcal{C}_{e_{a_1}}, q) = \text{exp}(0.3+0.1)/(\text{exp}(0.3+0.1)+\text{exp}(0.6)) = 0.45$. $P(a_2|\mathcal{C}_{e_{a_2}}, q) = \text{exp}(0.6)/(\text{exp}(0.3+0.1)+\text{exp}(0.6)) = 0.55$.}
%     \label{fig: concept regularize}
% \end{figure}
Assume we have an LP query $q=(e', r_q, ?, t_q)$, and at search step $l$, we have an action probability from policy $\pi(a|s^{(l)})$ for each candidate action $a \in \mathcal{A}^{(l)}$. We collect the concepts $\mathcal{C}_{e_a}$ of $e_a$ in each action $a$ and compute a concept-aware action probability
\begin{figure}[htbp]
    \centering
    \includegraphics[width=0.9\textwidth]{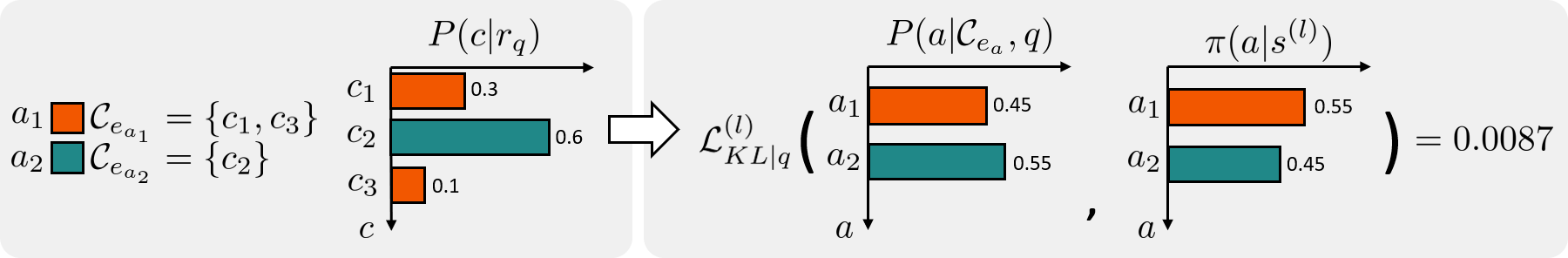}
    \caption{Concept regularizer. $P(a_1|\mathcal{C}_{e_{a_1}}, q) = \text{exp}(0.3+0.1)/(\text{exp}(0.3+0.1)+\text{exp}(0.6)) = 0.45$. $P(a_2|\mathcal{C}_{e_{a_2}}, q) = \text{exp}(0.6)/(\text{exp}(0.3+0.1)+\text{exp}(0.6)) = 0.55$.}
    \label{fig: concept regularize}
\end{figure}
\begin{equation}
    P(a|\mathcal{C}_{e_a}, q) = \frac{\text{exp}(\sum_{c\in \mathcal{C}_{e_a}}P(c|r_q))}{\sum_{a'\in \mathcal{A}^{(l)}}\text{exp}(\sum_{c'\in \mathcal{C}_{e_{a'}}}P(c'|r_q))}
\end{equation}
We then compute the Kullback-Leibler (KL) divergence between $P(a|\mathcal{C}_{e_a},q)$ and $\pi(a|s^{(l)})$ and minimize it during parameter optimization.
\begin{equation}
\label{eq: kl}
    \mathcal{L}_{\text{KL}|q}^{(l)} = \sum_{a\in \mathcal{A}^{(l)}} \pi(a|s^{(l)}) \log \left(\frac{\pi(a|s^{(l)})}{P(a|\mathcal{C}_{e_a}, q)}\right).
\end{equation}
Note that $r_q \in \mathcal{R}$ is observable in $\mathcal{G}_{\text{back}}$. $\mathcal{G}_{\text{back}}$ is huge and contains a substantial number of facts of $r_q$. As stated in FILT \cite{ding2022few}, although we have only $K$ associated edges for each unseen entity $e'$, its concepts $\mathcal{C}_{e'}$ is known. Our concept regularizer enables a parameter-free approach to match the concept-aware action probability $P(a|\mathcal{C}_{e_a},q)$ with the action probability taken from the policy $\pi(a|s^{(l)})$. It can be taken as guiding the policy to conform to the distribution of $r_q$'s objects' concepts observed in $\mathcal{G}_{\text{back}}$. We illustrate our concept regularizer in Fig. \ref{fig: concept regularize}.
% choose the action that goes to the entity whose concepts best fit $r_q$'s objects observed in $\mathcal{G}_{\text{back}}$.
% Note that although we are doing LP over TKG facts concerning unseen entities, $P(c|r)$ can still be extracted from $\mathcal{G}_{\text{back}}$ because $\mathcal{G}_{\text{back}}$ is huge and contains a substantial number of associated facts for every seen entities.
% sum up the number of appearance $n_{c|\mathcal{A}^{(l)}}$ of each concept $c$. We compute a probability $P(c|r_q) = $ denoting how probable it is to see $c$ if we follow the policy $\pi(a|s^{(l)})$.
% \begin{equation}
%     P(c|\pi(a|s^{(l)})) = \sum_{c_{e_a}}\left(\pi(a|s^{(l)}) \frac{n_{c}}{\sum_{c\in \mathcal{C}}n_{c}}\right)
% \end{equation}

\subsection{Parameter Learning}
Following \cite{ding2022few}, we train FITCARL with episodic training. In each episode, a training task $T$ is sampled, where we sample a $Sup_{e'}$ for every unseen entity $e' \in \mathcal{E}'_{\text{meta-train}}$ ($\mathcal{E}_T = \mathcal{E}'_{\text{meta-train}}$) and calculate loss over $Que_{e'}$. For each LP query $q$, we aim to maximize the cumulative reward along $L$ steps of search. We write our loss function (we minimize our loss) for each training task $T$ as follows.
\begin{equation}
\label{eq: loss}
\resizebox{0.95\columnwidth}{!}{%
$
\begin{aligned}
    &\mathcal{L}_T = \frac{1}{\sum_{e'} |Que_{e'}|}\sum_{e'} \sum_{q \in Que_{e'}} \sum_{l=0}^{L-1}
    \gamma^l \mathcal{L}_q^{(l)}, \ \  \mathcal{L}_q^{(l)} = \eta \mathcal{L}_{\text{KL}|q}^{(l)} -
    % \sum_{z=0}^l 
    \log(\pi(a^{(l)}|s^{(l)}))R(s^{(l)}, a^{(l)}).\\
    % \left(\pi(a^{(l)}|s^{(l)})R(s^{(l)}, a^{(l)}) + \eta \mathcal{L}_{\text{KL}|q}^{(l)}\right),\\
    % & \mathcal{L}_q^{(l)} = \pi(a^{(l)}|s^{(l)})R(s^{(l)}, a^{(l)}) + \eta \mathcal{L}_{\text{KL}|q}^{(l)}.
\end{aligned}
$
}
\end{equation}
$a^{(l)}$ is the selected action at search step $l$. $\gamma^l$ is the $l^{\text{th}}$ order of a 
discount factor $\gamma \in [0,1)$. $\eta$ is a hyperparameter deciding the magnitude of concept regularization. We use Algorithm 1 in Appendix E to further illustrate our meta-training process.
\section{Experiments}
We compare FITCARL with baselines on TKG few-shot OOG LP (Section \ref{sec: main results}). In Section \ref{sec: ablation study}, we first do several ablation studies to study the effectiveness of different model components. We then plot the performance over time to show FITCARL's robustness and present a case study to show FITCARL's explainability and the importance of learning confidence. We provide implementation details in Appendix A.
\subsection{Experimental Setting}
% \paragraph{Datasets}
We do experiments on three datasets proposed in \cite{ding2022few}, i.e., ICEWS14-OOG, ICEWS18-OOG and ICEWS0515-OOG. They contain the timestamped political facts in 2014, 2018 and from 2005 to 2015, respectively. All of them are constructed by taking the facts from the ICEWS \cite{DVN/28075_2015} TKB. Dataset statistics are shown in Table \ref{tab: data statistics}. We employ two evaluation metrics, i.e., mean reciprocal rank (MRR) and Hits@1/3/10. 
We provide detailed definitions of both metrics in Appendix D. We use the filtered setting proposed in \cite{DBLP:conf/nips/BordesUGWY13} for fairer evaluation.
% For every LP query $q$, we compute the rank $rank_q$ of the ground truth missing entity. MRR is defined as: $\frac{1}{\sum_{e' \in \mathcal{E}'_{\text{meta-test}}} |Que_{e'}|}\sum_{e' \in \mathcal{E}'_{\text{meta-test}}} \sum_{q \in Que_{e'}} \frac{1}{rank_q}$. Hits@1/3/10 denote the proportions of the predicted links where ground truth missing entities are ranked as top 1, top3, top10, respectively. We also use the filtered setting proposed in \cite{DBLP:conf/nips/BordesUGWY13} for fairer evaluation.
% \paragraph{Evaluation Protocol}
% We use two evaluation metrics, i.e., mean reciprocal rank (MRR) and Hits@1/3/10. For every LP query $q$, we compute the rank $\theta_q$ of the ground truth missing entity. MRR is defined as: $\frac{1}{\sum_{e' \in \mathcal{E}'_{\text{meta-test}}} |Que_{e'}|}\sum_{e' \in \mathcal{E}'_{\text{meta-test}}}\\ \sum_{q \in Que_{e'}} \frac{1}{\theta_q}$. Hits@1/3/10 denote the proportions of the predicted links where ground truth missing entities are ranked as top 1, top3, top10, respectively.
% \paragraph{Baselines}
For baselines, we consider the following methods. (1) Two traditional KGC methods, i.e., ComplEx \cite{DBLP:conf/icml/TrouillonWRGB16} and BiQUE \cite{DBLP:conf/emnlp/GuoK21}. (2) Three traditional TKGC methods, i.e., TNTComplEx \cite{DBLP:conf/iclr/LacroixOU20}, TeLM \cite{DBLP:conf/naacl/XuCNL21}, and TeRo \cite{DBLP:conf/coling/XuNAYL20}. (3) Three inductive KGC methods, i.e., MEAN \cite{DBLP:conf/ijcai/HamaguchiOSM17}, LAN \cite{DBLP:conf/aaai/WangHLP19}, and GEN \cite{DBLP:conf/nips/BaekLH20}. Among them, only GEN is trained with a meta-learning framework. (4) Two inductive TKG reasoning methods, including an inductive TKG forecasting method TITer \cite{DBLP:conf/emnlp/SunZMH021}, and a meta-learning-based inductive TKGC method FILT \cite{ding2022few} (FILT is the only previous work developed to solve TKG few-shot OOG LP). We take the experimental results of all baselines (except TITer) from \cite{ding2022few}. Following \cite{ding2022few}, we train TITer over all the TKG facts in $\mathcal{G}_{\text{back}}$ and $\mathbb{T}_{\text{meta-train}}$. We constrain TITer to only observe support quadruples of each test entity in $\mathcal{E}'_{\text{meta-test}}$ for inductive learning during inference. All methods are tested over exactly the same test examples.
\begin{table}[htbp]
    \centering
    \caption{Dataset statistics. 
    % $|\mathcal{E}'_{\text{meta-train}}|$, $|\mathcal{E}'_{\text{meta-valid}}|$, $|\mathcal{E}'_{\text{meta-test}}|$ are unseen entity numbers for meta-training, meta-validation, meta-test set, respectively. 
    % $N_{\text{back}}$ is the number of quadruples in $\mathcal{G}_{\text{back}}$. $N_{\text{meta-train}}$, $N_{\text{meta-valid}}$, $N_{\text{meta-test}}$ are quadruples numbers containing in $\mathbb{T}_{\text{meta-train}}$, $\mathbb{T}_{\text{meta-valid}}$, $\mathbb{T}_{\text{meta-test}}$, respectively.}
    }
    \resizebox{0.9\columnwidth}{!}{
%\centering
%\setlength{\tabcolsep}{1mm}
\begin{tabular}{c c c c c c c c c c c} \hline
\textbf{Dataset}&$|\mathcal{E}|$&$|\mathcal{R}|$&$|\mathcal{T}|$&$|\mathcal{E}'_{\text{meta-train}}|$&$|\mathcal{E}'_{\text{meta-valid}}|$&$|\mathcal{E}'_{\text{meta-test}}|$&$|\mathcal{G}_{\text{back}}|$&$|\mathbb{T}_{\text{meta-train}}|$&$|\mathbb{T}_{\text{meta-valid}}|$&$|\mathbb{T}_{\text{meta-test}}|$\\ 
\hline
ICEWS14-OOG  & 7128 & 230 & 365 & 385 & 48 & 49 & 83448 & 5772 & 718 & 705 \\ 
ICEWS18-OOG  & 23033 & 256 & 304 & 1268 & 160 & 158 & 444269 & 19291 & 2425 & 2373\\ 
ICEWS0515-OOG  & 10488 & 251 & 4017 & 647 & 80 & 82 & 448695 & 10115 & 1217 & 1228\\ 
\hline
\end{tabular}
}
\label{tab: data statistics}
\end{table}
\subsection{Main Results}
\label{sec: main results}
Table \ref{tab: main results} shows the experimental results of TKG 1-shot/3-shot OOG LP. We observe that traditional KGC and TKGC methods are beaten by inductive learning methods. It is because traditional methods cannot handle unseen entities. Besides, we also find that meta-learning-based methods, i.e., GEN, FILT and FITCARL, show better performance than other inductive learning methods. This is because meta-learning is more suitable for dealing with few-shot learning problems. 
% MEAN and LAN require a big size of auxiliary set for inference, however, the size of it is cut to 1 or 3 in the 1-shot or 3-shot setting. This makes them fail to optimally learn inductive representation of unseen entities. 
FITCARL shows superior performance over all metrics on all datasets. It outperforms the previous stat-of-the-art FILT with a huge margin. 
We attribute it to several reasons. (1) Unlike FILT that uses KG score function over all the entities for prediction, FITCARL is an RL-based method that directly searches the predicted answer through their multi-hop temporal neighborhood, making it better capture highly-related graph information through time. (2) FITCARL takes advantage of its confidence learner. It helps to alleviate the negative impact from the few-shot setting. (3) Concept regularizer serves as a strong tool for exploiting concept-aware information in TKBs and adaptively guides FITCARL to learn a policy that conforms to the concept distribution shown in $\mathcal{G}_{\text{back}}$.
\begin{table*}[htbp]
\caption{Experimental results of TKG 1-shot and 3-shot OOG LP. Evaluation metrics are MRR and Hits@1/3/10 (H@1/3/10).
    Best results are marked bold.}\label{tab: main results}
    \centering
    \resizebox{\columnwidth}{!}{
    \begin{tabular}{@{}lcccc cccc cccc cccc cccc cccc@{}}
\toprule
        \textbf{Datasets} & \multicolumn{8}{c}{\textbf{ICEWS14-OOG}} & \multicolumn{8}{c}{\textbf{ICEWS18-OOG}} & \multicolumn{8}{c}{\textbf{ICEWS0515-OOG}}\\
\cmidrule(lr){2-9} \cmidrule(lr){10-17} \cmidrule(lr){18-25}
& \multicolumn{2}{c}{MRR} & \multicolumn{2}{c}{H@1} & \multicolumn{2}{c}{H@3} & \multicolumn{2}{c}{H@10}& \multicolumn{2}{c}{MRR} & \multicolumn{2}{c}{H@1} & \multicolumn{2}{c}{H@3} & \multicolumn{2}{c}{H@10} & \multicolumn{2}{c}{MRR} & \multicolumn{2}{c}{H@1} & \multicolumn{2}{c}{H@3} & \multicolumn{2}{c}{H@10}\\

 \textbf{Model}& 1-S & 3-S & 1-S & 3-S & 1-S & 3-S & 1-S & 3-S & 1-S & 3-S & 1-S & 3-S & 1-S & 3-S & 1-S & 3-S & 1-S & 3-S & 1-S & 3-S & 1-S & 3-S & 1-S & 3-S\\
\midrule
        ComplEx & .048 & .046& .018 & .014&.045 & .046&.099 & .089
        & .039 & .044& .031 & .026&.048 & .042& .085 &.093
        & .077 & .076&.045 & .048&.074 & .071&.129 &.120
         \\
        BiQUE & .039 & .035& .015 & .014& .041 & .030& .073 &.066
        & .029 & .032& .022 & .021& .033 & .037& .064 &.073
        & .075 & .083&.044 & .049&.072 & .077&.130 &.144
         \\
\midrule 
        % TARGCN & 8.09 & 6.42 & 8.16 & 8.27 & 10.57 
        % & 0.79 & 0.23 & 0.49 & 0.81 & 1.38
        %  \\
        TNTComplEx & .043 & .044&.015 & .016&.033 & .042&.102&.096 
        & .046 & .048&.023 & .026&.043 & .044&.087& .082
        & .034 & .037&.014 & .012&.031 & .036&.060& .071
         \\
        TeLM & .032 & .035&.012 & .009&.021 & .023&.063 &.077
        & .049 & .019&.029 & .001&.045 & .013&.084&.054
        & .080 & .072&.041 & .034&.077 & .072&.138&.151
         \\
        TeRo  & .009 & .010&.002 & .002&.005 & .002&.015&.020 
        & .007 & .006&.003 & .001&.006 & .003&.013&.006
        & .012 & .023&.000 & .010&.008 & .017&.024&.040
         \\
        % TITer  & .144 & .200&.105 & .148&.163 & .226&.228&.314 
        % & .064 & .115&.038 & .076&.075 & .131&.011&.186
        % & .115 & .228&.080 & .168&.130 & .262&.173&.331
        %  \\
        % TeRO  & 35.38 & 24.42 & 39.21 & 47.74 & 59.12 
        % & 8.28 & 4.74 & 8.36 & 10.66 & 14.50
        
\midrule
        MEAN  & .035 & .144&.013 & .054&.032 & .145&.082&.339 
        & .016 & .101&.003 & .014&.012 & .114&.043&.283
        & .019 & .148&.003 & .039&.017 & .175&.052&.384
       \\
       LAN  & .168 & .199&.050 & .061&.199 & .255&.421&.500 
        & .077 & .127&.018 & .025&.067 & .165&.199&.344
        & .171 & .182&.081 & .068&.180 & .191&.367&.467
       \\
        %  VN Network & - & - & - & - 
        % & - & - & - & -
        % & - & - & - & -
        %  \\
% \midrule     
      GEN  & .231 & .234&.162 & .155&.250 & .284&.378&.389 
        & .171 & .216&.112 & .137&.189 & .252&.289&.351
        & .268 & .322&.185 & .231&.308 & .362&.413&.507
        \\       
\midrule
        TITer  & .144 & .200&.105 & .148&.163 & .226&.228&.314 
        & .064 & .115&.038 & .076&.075 & .131&.011&.186
        & .115 & .228&.080 & .168&.130 & .262&.173&.331
         \\
        FILT 
        & .278 & .321&.208 & .240&.305 &.357 &.410&.475
        & .191 & .266&.129 & .187&.209 & .298&.316&.417
        & .273 & .370&.201 & .299&.303 & .391&.405&.516
        \\
        FITCARL
        & \textbf{.418} & \textbf{.481}&\textbf{.284} & \textbf{.329}&\textbf{.522} &\textbf{.646} &\textbf{.681}&\textbf{.696}
        & \textbf{.297} & \textbf{.370}&\textbf{.156} & \textbf{.193}&\textbf{.386} & \textbf{.559}&\textbf{.584}&\textbf{.627}
        & \textbf{.345} & \textbf{.513}&\textbf{.202} & \textbf{.386}&\textbf{.482} & \textbf{.618}&\textbf{.732}&\textbf{.700}
        \\
\bottomrule
    \end{tabular}
    }
\end{table*}
\subsection{Further Analysis}
\subsubsection{Ablation Study}
\label{sec: ablation study}
We conduct several ablation studies to study the effectiveness of different model components. \textbf{(A) Action Space Sampling Variants:} To prevent oversized action space $\mathcal{A}^{(l)}$, we use a time-adaptive sampling method (see Section \ref{sec: RL framework}). We show its effectiveness by switching it to random sample (ablation A1) and time-proximity sample (ablation A2). In time-proximity sample, we take a fixed number of outgoing edges temporally closest to the current node at $t^{(l)}$ as $\mathcal{A}^{(l)}$. We keep $|\mathcal{A}^{(l)}|$ unchanged. 
% From Table \ref{tab: ablation results 1shot} and \ref{tab: ablation results 3shot}, we observe that time-adaptive sample is effective. It helps FITCARL to sample the most temporally contributive actions along the whole time axis. 
\textbf{(B) Removing Confidence Learner:} In ablation B, we remove the confidence learner. 
% It is shown in Table \ref{tab: ablation results 1shot} and \ref{tab: ablation results 3shot} that incorporating confidence can greatly help FITCARL in action selection. 
\textbf{(C) Removing Concept Regularizer:} In ablation C, we remove concept regularizer. 
% We find that concept regularizer serves as a strong tool to guide FITCARL in prediction. 
\textbf{(D) Time-Aware Transformer Variants:} 
% In abaltion D1, we switch our time-aware transformer to mean aggregation, i.e., $\mathbf{h}_{e'|q} = 1/K \sum_{i=1}^K \mathbf{h}_{e'}^i$. 
% In ablation D2, 
We remove the time-aware positional encoding method by deleting the second term of Equation \ref{eq: time att transformer}.
% and use a time-unaware transformer for unseen entity learning.
\textbf{(E) Removing Temporal Reasoning Modules:} In ablation E, we study the importance of temporal reasoning. 
We first combine ablation A1 and D, and then delete every term related to time difference representations computed with Equation \ref{eq: time encoding}. We create a model variant without using any temporal information (see Appendix C for detailed setting). We present the experimental results of ablation studies in Table \ref{tab: ablation}. From ablation A1 and A2, we observe that time-adaptive sample is effective. 
% It helps FITCARL to sample the most temporally contributive actions along the whole time axis. 
We also see a great performance drop in ablation B and C, indicating the strong importance of our confidence learner and concept regularizer. We only do ablation D for 3-shot model because in 1-shot case our model does not need to distinguish the importance of multiple support quadruples. We find that our time-aware positional encoding makes great contribution. Finally, we observe that ablation E shows poor performance (worse than A1 and D in most cases), implying that incorporating temporal information is essential for FITCARL to solve TKG few-shot OOG LP.
\begin{table*}[htbp]
\caption{Ablation study results. Best results are marked bold.}\label{tab: ablation}
    \centering
    \resizebox{\columnwidth}{!}{
    \begin{tabular}{@{}lcccc cccc cccc cccc cccc cccc@{}}
\toprule
        \textbf{Datasets} & \multicolumn{8}{c}{\textbf{ICEWS14-OOG}} & \multicolumn{8}{c}{\textbf{ICEWS18-OOG}} & \multicolumn{8}{c}{\textbf{ICEWS0515-OOG}}\\
\cmidrule(lr){2-9} \cmidrule(lr){10-17} \cmidrule(lr){18-25}
& \multicolumn{2}{c}{MRR} & \multicolumn{2}{c}{H@1} & \multicolumn{2}{c}{H@3} & \multicolumn{2}{c}{H@10}& \multicolumn{2}{c}{MRR} & \multicolumn{2}{c}{H@1} & \multicolumn{2}{c}{H@3} & \multicolumn{2}{c}{H@10} & \multicolumn{2}{c}{MRR} & \multicolumn{2}{c}{H@1} & \multicolumn{2}{c}{H@3} & \multicolumn{2}{c}{H@10}\\

 \textbf{Model}& 1-S & 3-S & 1-S & 3-S & 1-S & 3-S & 1-S & 3-S & 1-S & 3-S & 1-S & 3-S & 1-S & 3-S & 1-S & 3-S & 1-S & 3-S & 1-S & 3-S & 1-S & 3-S & 1-S & 3-S\\
\midrule
        A1 & .404 & .418& .283 & .287&.477 & .494&.647 & .667
        & .218 & .260& .153 & .167&.220 & .296& .404 &.471
        & .190 & .401&.108 & .289&.196 & .467&.429 &.624
         \\
        A2 & .264 & .407& .241 & .277& .287 & .513& .288 &.639
        & .242 & .265& .126 & .168& .337 & .291& .444 &.499
        & .261 & .414&.200 & .267&.298 & .545&.387 &.640
         \\
\midrule 
        B & .373 & .379& .255 & .284& .454 & .425& .655 &.564
        & .156 & .258& .106 & .191& .162 & .271& .273 &.398
        & .285 & .411&.198 & .336&.328 & .442&.447 &.567
         \\
        % B & .373 & .044&.015 & .016&.033 & .042&.102&.096 
        % & .046 & .048&.023 & .026&.043 & .044&.087& .082
        % & .034 & .037&.014 & .012&.031 & .036&.060& .071
        %  \\
\midrule
        C & .379 & .410&.265 & .236&.489 & .570&.667 &.691
        & .275 & .339&.153 & .190&.346 & .437&.531&.556
        & .223 & .411&.130 & .243&.318 & .544&.397&.670
         \\
\midrule
        % D1  & - & .399&- & .275&- & .487&-&.628 
        % & - & .267&- & .193&- & .277&-&.454
        % & - & .351&- & .213&- & .431&-&.635
        %  \\

        D  & - & .438&- & .262&- & .626&-&.676 
        & - & .257&- & .160&- & .280&-&.500
        & - & .438&- & .262&- & .610&-&.672
         \\
\midrule
        E  & .270 & .346&.042 & .178&.480 & .466&.644&.662 
        & .155 & .201&.012 & .117&.197 & .214&.543&.429
        & .176 & .378&.047 & .239&.194 & .501&.506&.584
         \\
\midrule
        FITCARL
        & \textbf{.418} & \textbf{.481}&\textbf{.284} & \textbf{.329}&\textbf{.522} &\textbf{.646} &\textbf{.681}&\textbf{.696}
        & \textbf{.297} & \textbf{.370}&\textbf{.156} & \textbf{.193}&\textbf{.386} & \textbf{.559}&\textbf{.584}&\textbf{.627}
        & \textbf{.345} & \textbf{.513}&\textbf{.202} & \textbf{.386}&\textbf{.482} & \textbf{.618}&\textbf{.732}&\textbf{.700}
        \\
\bottomrule
    \end{tabular}
    }
\end{table*}
\subsubsection{Performance over Time}
\label{sec: performance over time}
\begin{figure*} 
    \centering
  \subfloat[\label{fig: ICEWS14-1}\small ICEWS14-OOG 1-shot]{%
       \includegraphics[width=0.33\textwidth]{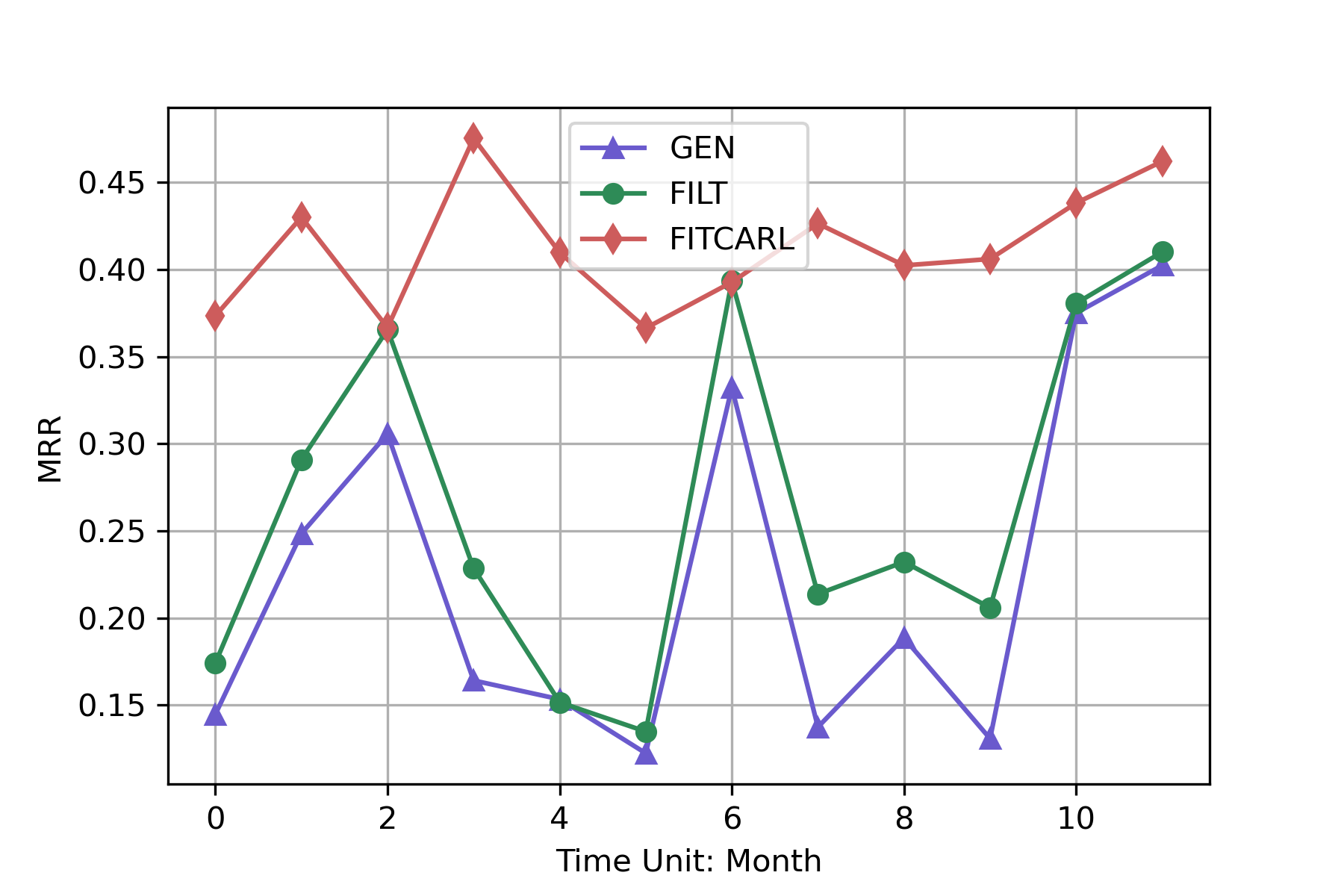}}
    \hfill
  \subfloat[\label{fig: ICEWS18-1}\small ICEWS18-OOG 1-shot]{%
        \includegraphics[width=0.33\textwidth]{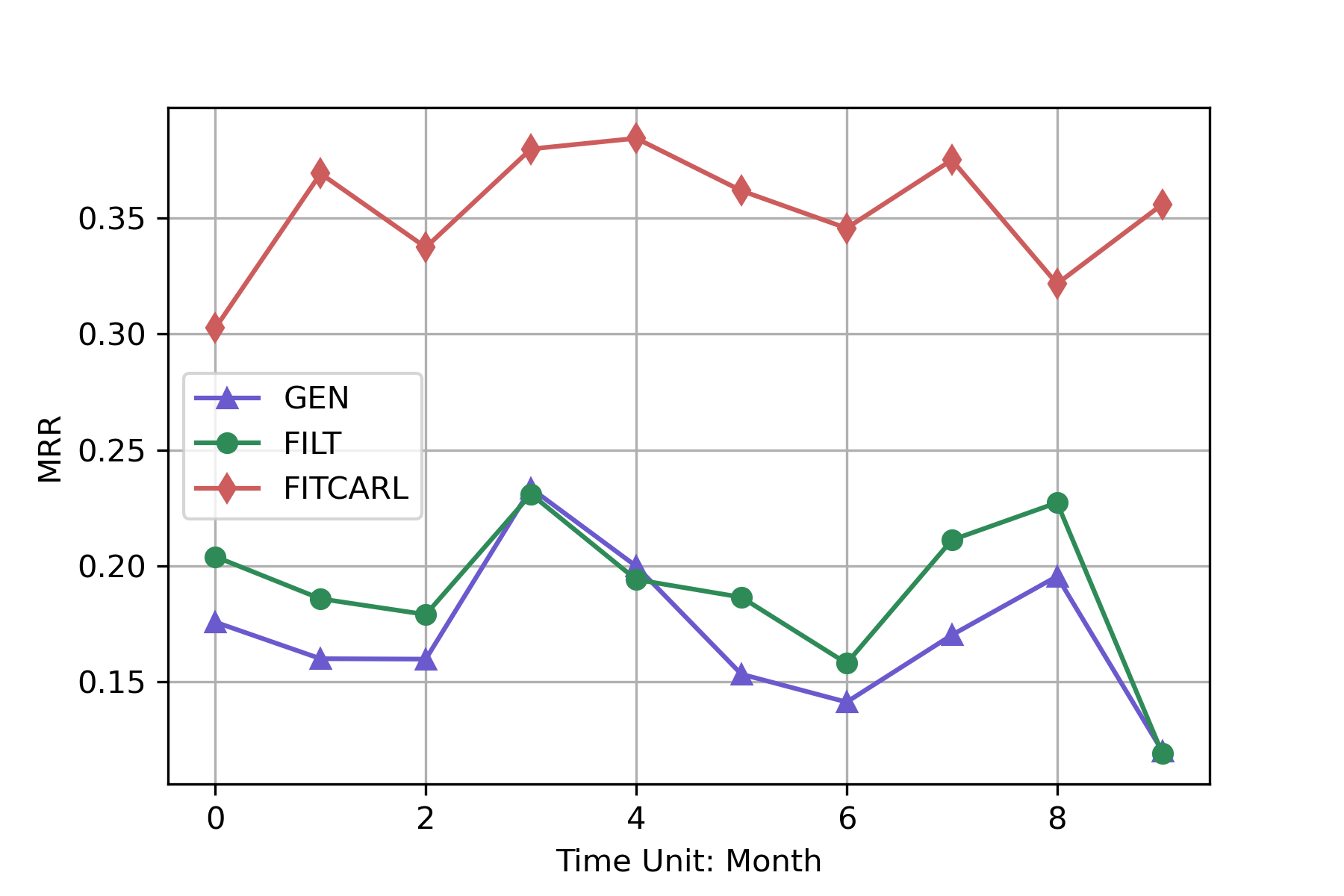}}
  \subfloat[\label{fig: ICEWS05-15-1}\small ICEWS0515-OOG 1-shot]{%
        \includegraphics[width=0.33\textwidth]{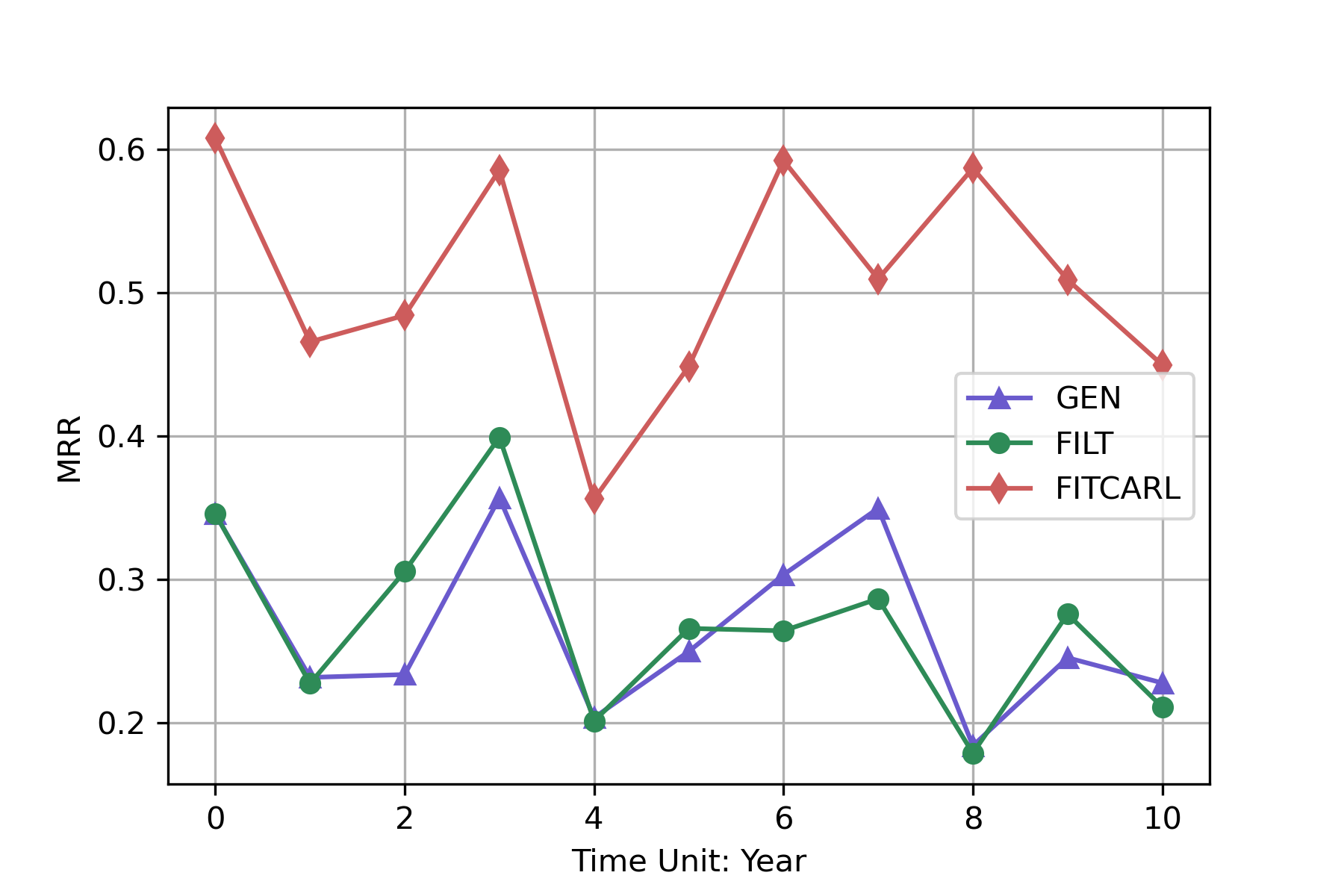}}
    \hfill
  \subfloat[\label{fig: ICEWS14-3}\small ICEWS14-OOG 3-shot]{%
        \includegraphics[width=0.33\textwidth]{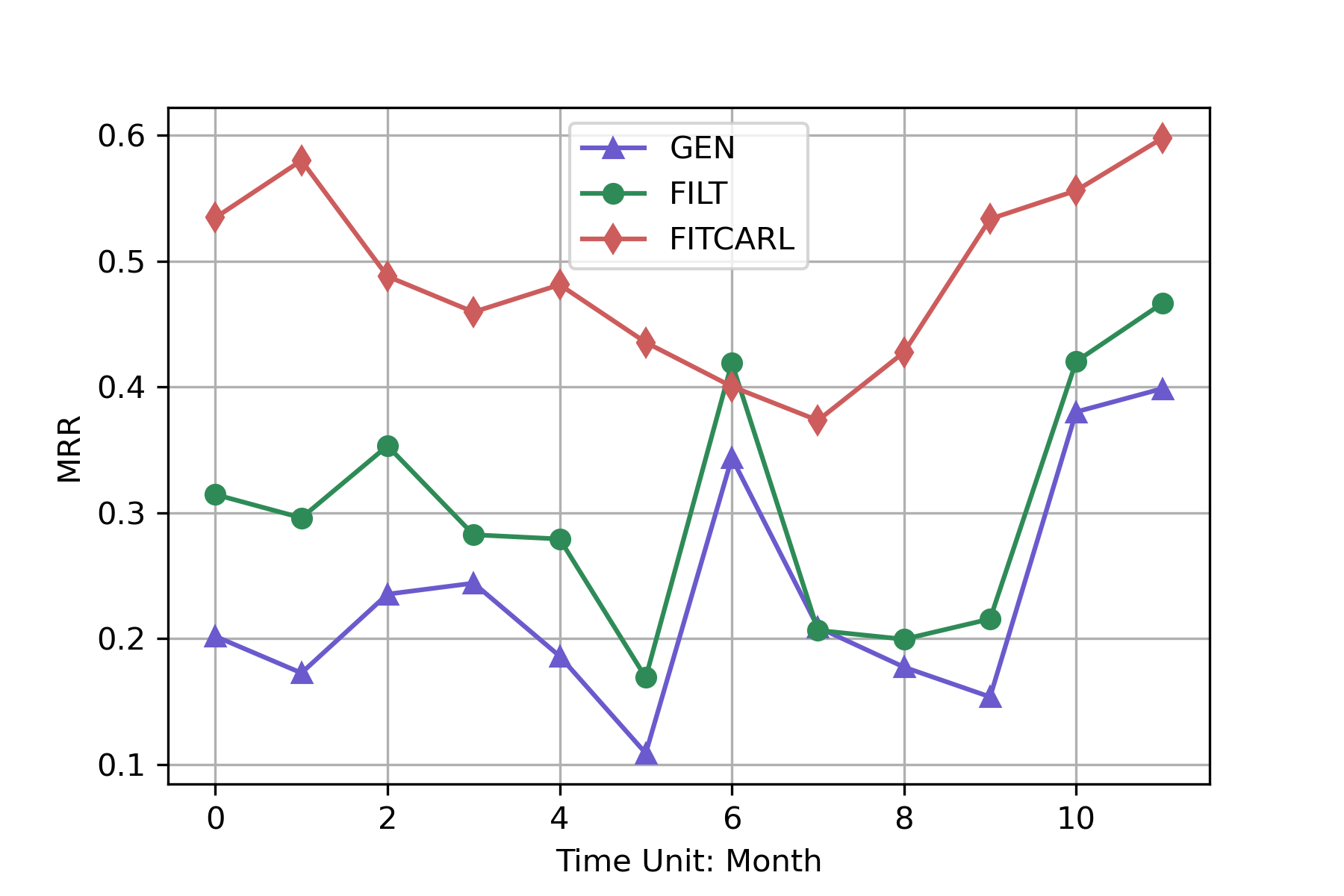}}
    \hfill
  \subfloat[\label{fig: ICEWS18-3}\small ICEWS18-OOG 3-shot]{%
        \includegraphics[width=0.33\textwidth]{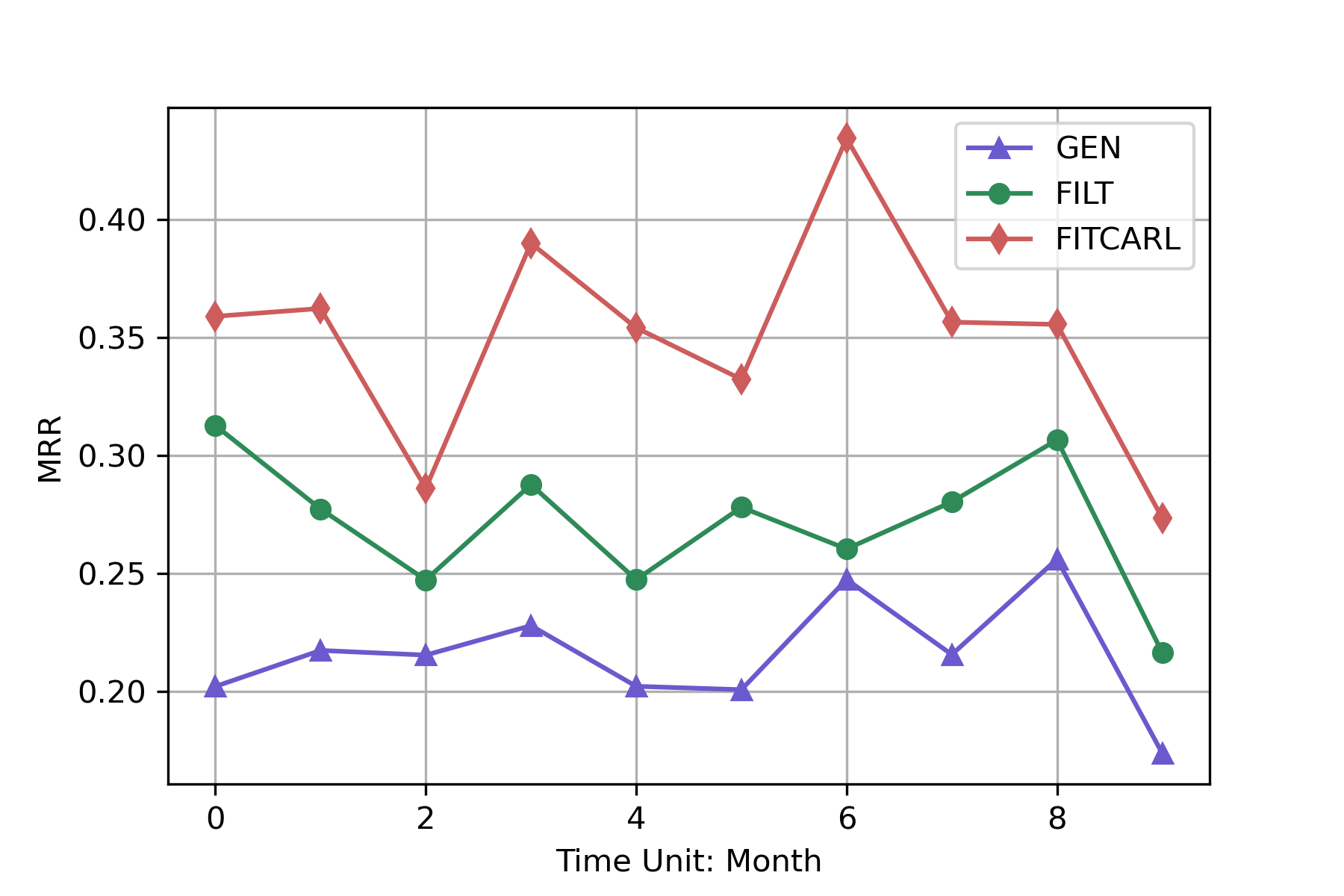}}
    \hfill
  \subfloat[\label{fig: ICEWS05-15-3}\small ICEWS0515-OOG 3-shot]{%
        \includegraphics[width=0.33\textwidth]{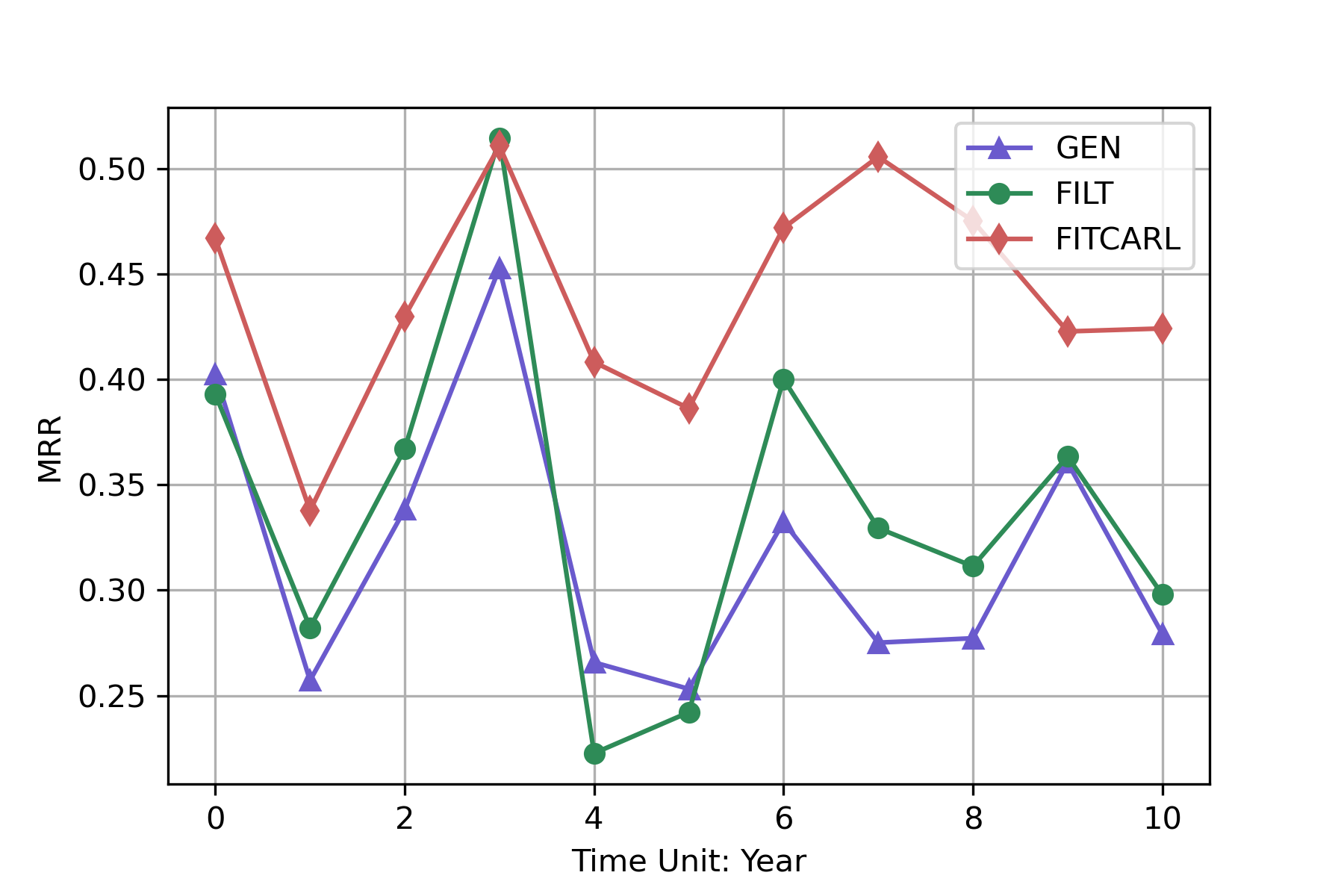}}
    \hfill
  \caption{Performance comparison among FITCARL, FILT and GEN over different query time $t_q$. Horizontal axis of each subfigure denotes how temporally faraway from the first timestamp. We aggregate the performance of each month to one point in ICEWS14-OOG and ICEWS18-OOG. A point for ICEWS0515-OOG denotes the aggregated performance in each year.}
  \label{fig: sup que diff} 
\end{figure*}

To demonstrate the robustness of FITCARL, we plot its MRR performance over prediction time (query time $t_q$). We compare FITCARL with two meta-learning-based strong baselines GEN and FILT. From Figure \ref{fig: ICEWS14-1} to \ref{fig: ICEWS05-15-3}, we find that our model can constantly outperform baselines. This indicates that FITCARL improves LP performance for examples existing at almost all timestamps, proving its robustness. GEN is not designed for TKG reasoning, and thus it cannot show optimal performance. Although FILT is designed for TKG few-shot OOG LP, we show that our RL-based model is much stronger. 

\subsubsection{Case Study}
\label{sec: case study}
\begin{figure*} 
    \centering
  \subfloat[\label{fig: case1_w}\small FITCARL]{%
       \includegraphics[width=0.47\textwidth]{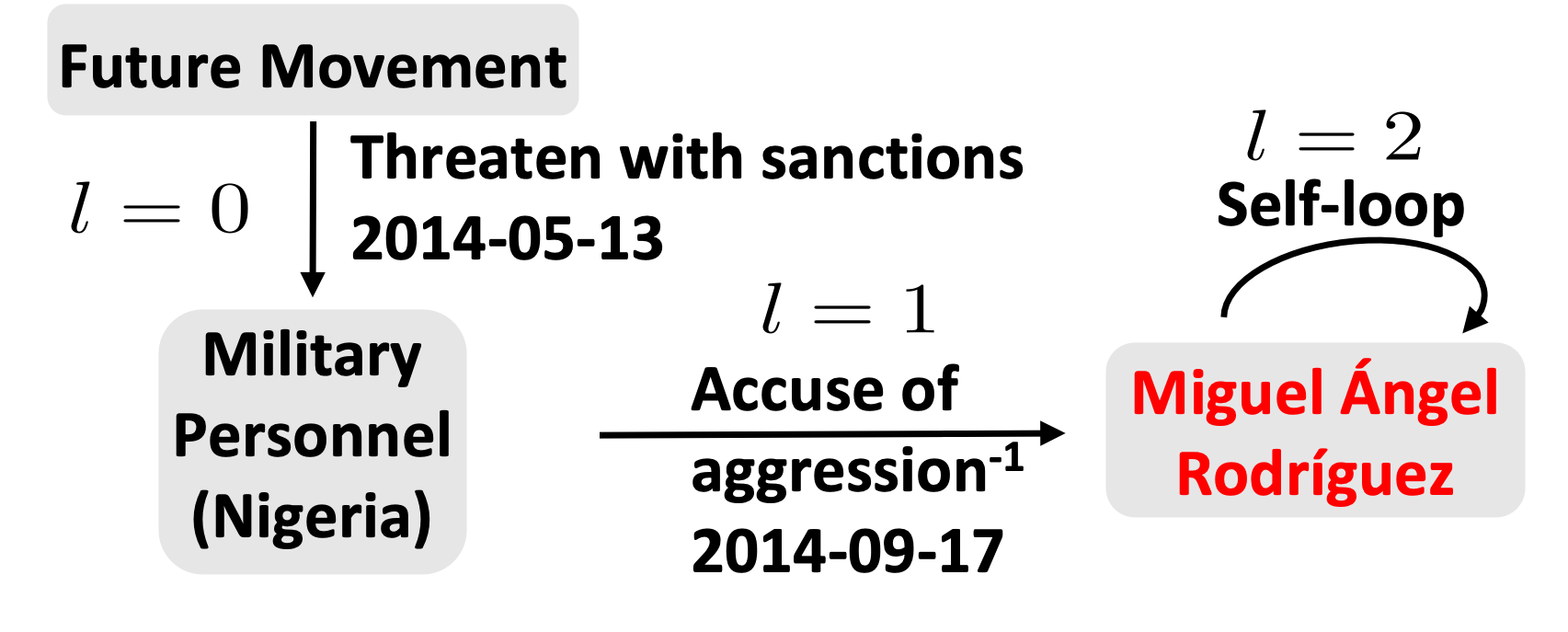}}
    \hfill
  \subfloat[\label{fig: case1_wo}\small FITCARL w.o. Confidence]{%
        \includegraphics[width=0.47\textwidth]{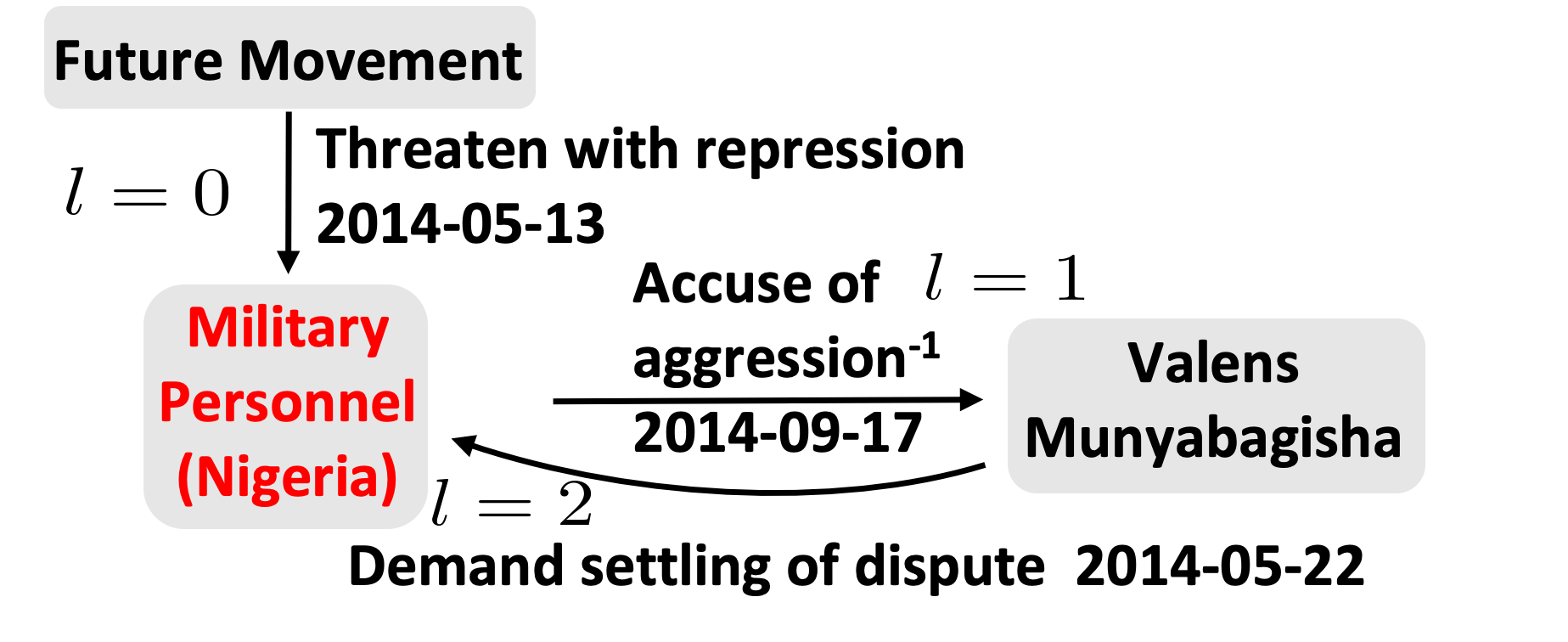}}
    \hfill
  \caption{Case study reasoning path visualization. The entity marked in red are the answer predicted by the model. w.o. means without.}
  \label{fig: case study} 
\end{figure*}

We do a case study to show how FITCARL provides explainability and how confidence learner helps in reasoning. 
% Each case corresponds to predicting a link derived from a query quadruple in ICEWS14-OOG. For each case, we visualize a specific reasoning path of FITCARL as well as its variant without the confidence learner. 
% Case 1 is presented in Fig. \ref{fig: case study}. 
% In Case 1, 
We ask 3-shot FITCARL and its variant without the confidence learner (both trained on ICEWS14-OOG) to predict the missing entity of the LP query (\textit{Future Movement}, \textit{Express intent to cooperate on intelligence}, $?$, 2014-11-12), where \textit{Future Movement} is a newly-emerged entity that is unseen during training and the answer to this LP query is \textit{Miguel Ángel Rodríguez}. We visualize a specific reasoning path of each model and present them in Fig. \ref{fig: case study}.
The relation \textit{Express intent to cooperate on intelligence} indicates a positive relationship between subject and object entities. FITCARL performs a search with length $L=3$, where it finds an entity \textit{Military Personnel (Nigeria)} that is in a negative relationship with both \textit{Future Movement} and \textit{Miguel Ángel Rodríguez}. FITCARL provides explanation by finding a reasoning path representing the proverb: The enemy of the enemy is my friend. For FITCARL without confidence learner, we find that it can also provide similar explanation by finding another entity that is also an enemy of \textit{Military Personnel (Nigeria)}. However, it fails to find the ground truth answer because it neglects the confidence of each action. The confidence learner assigns high probability to the ground truth entity, leading to a correct prediction.
% derived from the query quadruple (Future Movement, Express intent to cooperate on intelligence, Miguel Ángel Rodríguez, 2014-11-12).
% At the first search step, FITCARL starts from the newly-emerged entity and goes along the fact (\textit{Future Movement}, \textit{Threaten with sanctions}, \textit{Military Personnel (Nigeria)}, 2014-05-13) which indicates a negative relationship between its subject and object entities. Then, at step $2$, FITCARL goes along the fact (\textit{Military Personnel (Nigeria)}, \textit{Accuse of aggression}, \textit{Miguel Ángel Rodríguez}, 2014-05-13) which also indicates a negative relationship between both entities. FITCARL stops at \textit{Miguel Ángel Rodríguez} at the last search step.

\section{Conclusion}
We present an RL-based TKGC method FITCARL to solve TKG few-shot OOG LP, where models are asked to predict the links concerning newly-emerged entities that have only a few observed associated facts. FITCARL is a meta-learning-based model trained with episodic training. It learns representations of newly-emerged entities by using a time-aware Transformer. To further alleviate the negative impact of the few-shot setting, a confidence learner is proposed to be coupled with the policy network for making better decisions. A parameter-free concept regularizer is also developed to better exploit concept-aware information in TKBs. Experimental results show that FITCARL achieves a new state-of-the-art and provides explainability.
\bibliographystyle{splncs04}
\bibliography{sample}

\begin{thebibliography}{10}
\providecommand{\url}[1]{\texttt{#1}}
\providecommand{\urlprefix}{URL }
\providecommand{\doi}[1]{https://doi.org/#1}

\bibitem{DBLP:conf/nips/AbboudCLS20}
Abboud, R., Ceylan, {\.I}.{\.I}., Lukasiewicz, T., Salvatori, T.: Boxe: {A} box
  embedding model for knowledge base completion. In: NeurIPS (2020)

\bibitem{DBLP:conf/iclr/AmmanabroluH20}
Ammanabrolu, P., Hausknecht, M.J.: Graph constrained reinforcement learning for
  natural language action spaces. In: {ICLR}. OpenReview.net (2020)

\bibitem{DBLP:conf/nips/BaekLH20}
Baek, J., Lee, D.B., Hwang, S.J.: Learning to extrapolate knowledge:
  Transductive few-shot out-of-graph link prediction. In: NeurIPS (2020)

\bibitem{DBLP:conf/emnlp/BalazevicAH19}
Balazevic, I., Allen, C., Hospedales, T.M.: Tucker: Tensor factorization for
  knowledge graph completion. In: {EMNLP/IJCNLP} {(1)}. pp. 5184--5193.
  Association for Computational Linguistics (2019)

\bibitem{DBLP:conf/nips/BordesUGWY13}
Bordes, A., Usunier, N., Garc{\'{\i}}a{-}Dur{\'{a}}n, A., Weston, J.,
  Yakhnenko, O.: Translating embeddings for modeling multi-relational data. In:
  {NIPS}. pp. 2787--2795 (2013)

\bibitem{DVN/28075_2015}
Boschee, E., Lautenschlager, J., O'Brien, S., Shellman, S., Starz, J., Ward,
  M.: {ICEWS Coded Event Data} (2015)

\bibitem{DBLP:conf/acl/ChenWLL22}
Chen, K., Wang, Y., Li, Y., Li, A.: Rotateqvs: Representing temporal
  information as rotations in quaternion vector space for temporal knowledge
  graph completion. In: {ACL} {(1)}. pp. 5843--5857. Association for
  Computational Linguistics (2022)

\bibitem{DBLP:conf/emnlp/ChenZZCC19}
Chen, M., Zhang, W., Zhang, W., Chen, Q., Chen, H.: Meta relational learning
  for few-shot link prediction in knowledge graphs. In: {EMNLP/IJCNLP} {(1)}.
  pp. 4216--4225. Association for Computational Linguistics (2019)

\bibitem{DBLP:conf/emnlp/ChoMGBBSB14}
Cho, K., van Merrienboer, B., G{\"{u}}l{\c{c}}ehre, {\c{C}}., Bahdanau, D.,
  Bougares, F., Schwenk, H., Bengio, Y.: Learning phrase representations using
  {RNN} encoder-decoder for statistical machine translation. In: {EMNLP}. pp.
  1724--1734. {ACL} (2014)

\bibitem{DBLP:journals/corr/abs-2205-10621}
Ding, Z., He, B., Ma, Y., Han, Z., Tresp, V.: Learning meta representations of
  one-shot relations for temporal knowledge graph link prediction. CoRR
  \textbf{abs/2205.10621} (2022)

\bibitem{ding2022simple}
Ding, Z., Ma, Y., He, B., Han, Z., Tresp, V.: A simple but powerful graph
  encoder for temporal knowledge graph completion. In: NeurIPS 2022 Temporal
  Graph Learning Workshop

\bibitem{DBLP:journals/corr/abs-2208-06501}
Ding, Z., Qi, R., Li, Z., He, B., Wu, J., Ma, Y., Meng, Z., Han, Z., Tresp, V.:
  Forecasting question answering over temporal knowledge graphs. CoRR
  \textbf{abs/2208.06501} (2022)

\bibitem{ding2022few}
Ding, Z., Wu, J., He, B., Ma, Y., Han, Z., Tresp, V.: Few-shot inductive
  learning on temporal knowledge graphs using concept-aware information. In:
  4th Conference on Automated Knowledge Base Construction (2022)

\bibitem{DBLP:conf/emnlp/GuoK21}
Guo, J., Kok, S.: Bique: Biquaternionic embeddings of knowledge graphs. In:
  {EMNLP} {(1)}. pp. 8338--8351. Association for Computational Linguistics
  (2021)

\bibitem{DBLP:conf/ijcai/HamaguchiOSM17}
Hamaguchi, T., Oiwa, H., Shimbo, M., Matsumoto, Y.: Knowledge transfer for
  out-of-knowledge-base entities : {A} graph neural network approach. In:
  {IJCAI}. pp. 1802--1808. ijcai.org (2017)

\bibitem{DBLP:conf/emnlp/HanDMGT21}
Han, Z., Ding, Z., Ma, Y., Gu, Y., Tresp, V.: Learning neural ordinary
  equations for forecasting future links on temporal knowledge graphs. In:
  {EMNLP} {(1)}. pp. 8352--8364. Association for Computational Linguistics
  (2021)

\bibitem{DBLP:conf/cikm/HeWZTR20}
He, Y., Wang, Z., Zhang, P., Tu, Z., Ren, Z.: {VN} network: Embedding newly
  emerging entities with virtual neighbors. In: {CIKM}. pp. 505--514. {ACM}
  (2020)

\bibitem{DBLP:journals/corr/abs-1904-05530}
Jin, W., Zhang, C., Szekely, P.A., Ren, X.: Recurrent event network for
  reasoning over temporal knowledge graphs. CoRR  \textbf{abs/1904.05530}
  (2019)

\bibitem{DBLP:conf/kdd/JungJK21}
Jung, J., Jung, J., Kang, U.: Learning to walk across time for interpretable
  temporal knowledge graph completion. In: {KDD}. pp. 786--795. {ACM} (2021)

\bibitem{DBLP:conf/iclr/LacroixOU20}
Lacroix, T., Obozinski, G., Usunier, N.: Tensor decompositions for temporal
  knowledge base completion. In: {ICLR}. OpenReview.net (2020)

\bibitem{DBLP:conf/www/LeblayC18}
Leblay, J., Chekol, M.W.: Deriving validity time in knowledge graph. In: {WWW}
  (Companion Volume). pp. 1771--1776. {ACM} (2018)

\bibitem{DBLP:conf/acl/LiTZWYW21}
Li, J., Tang, T., Zhao, W.X., Wei, Z., Yuan, N.J., Wen, J.: Few-shot knowledge
  graph-to-text generation with pretrained language models. In: {ACL/IJCNLP}
  (Findings). Findings of {ACL}, vol. {ACL/IJCNLP} 2021, pp. 1558--1568.
  Association for Computational Linguistics (2021)

\bibitem{DBLP:conf/acl/LiJGLGWC20}
Li, Z., Jin, X., Guan, S., Li, W., Guo, J., Wang, Y., Cheng, X.: Search from
  history and reason for future: Two-stage reasoning on temporal knowledge
  graphs. In: {ACL/IJCNLP} {(1)}. pp. 4732--4743. Association for Computational
  Linguistics (2021)

\bibitem{DBLP:conf/aaai/LinLSLZ15}
Lin, Y., Liu, Z., Sun, M., Liu, Y., Zhu, X.: Learning entity and relation
  embeddings for knowledge graph completion. In: {AAAI}. pp. 2181--2187. {AAAI}
  Press (2015)

\bibitem{DBLP:conf/aaai/MessnerAC22}
Messner, J., Abboud, R., Ceylan, {\.I}.{\.I}.: Temporal knowledge graph
  completion using box embeddings. In: {AAAI}. pp. 7779--7787. {AAAI} Press
  (2022)

\bibitem{mirtaheri2021one}
Mirtaheri, M., Rostami, M., Ren, X., Morstatter, F., Galstyan, A.: One-shot
  learning for temporal knowledge graphs. In: 3rd Conference on Automated
  Knowledge Base Construction (2021)

\bibitem{DBLP:conf/icml/NickelTK11}
Nickel, M., Tresp, V., Kriegel, H.: A three-way model for collective learning
  on multi-relational data. In: {ICML}. pp. 809--816. Omnipress (2011)

\bibitem{DBLP:conf/nips/PaszkeGMLBCKLGA19}
Paszke, A., Gross, S., Massa, F., Lerer, A., Bradbury, J., Chanan, G., Killeen,
  T., Lin, Z., Gimelshein, N., Antiga, L., Desmaison, A., K{\"{o}}pf, A., Yang,
  E.Z., DeVito, Z., Raison, M., Tejani, A., Chilamkurthy, S., Steiner, B.,
  Fang, L., Bai, J., Chintala, S.: Pytorch: An imperative style,
  high-performance deep learning library. In: NeurIPS. pp. 8024--8035 (2019)

\bibitem{DBLP:conf/aaai/SadeghianACW21}
Sadeghian, A., Armandpour, M., Colas, A., Wang, D.Z.: Chronor: Rotation based
  temporal knowledge graph embedding. In: {AAAI}. pp. 6471--6479. {AAAI} Press
  (2021)

\bibitem{DBLP:conf/acl/SaxenaTT20}
Saxena, A., Tripathi, A., Talukdar, P.P.: Improving multi-hop question
  answering over knowledge graphs using knowledge base embeddings. In: {ACL}.
  pp. 4498--4507. Association for Computational Linguistics (2020)

\bibitem{DBLP:conf/esws/SchlichtkrullKB18}
Schlichtkrull, M.S., Kipf, T.N., Bloem, P., van~den Berg, R., Titov, I.,
  Welling, M.: Modeling relational data with graph convolutional networks. In:
  {ESWC}. Lecture Notes in Computer Science, vol. 10843, pp. 593--607. Springer
  (2018)

\bibitem{DBLP:conf/emnlp/ShengGCYWLX20}
Sheng, J., Guo, S., Chen, Z., Yue, J., Wang, L., Liu, T., Xu, H.: Adaptive
  attentional network for few-shot knowledge graph completion. In: {EMNLP}
  {(1)}. pp. 1681--1691. Association for Computational Linguistics (2020)

\bibitem{DBLP:conf/emnlp/SunZMH021}
Sun, H., Zhong, J., Ma, Y., Han, Z., He, K.: Timetraveler: Reinforcement
  learning for temporal knowledge graph forecasting. In: {EMNLP} {(1)}. pp.
  8306--8319. Association for Computational Linguistics (2021)

\bibitem{tresp2015learning}
Tresp, V., Esteban, C., Yang, Y., Baier, S., Krompa{\ss}, D.: Learning with
  memory embeddings. arXiv preprint arXiv:1511.07972  (2015)

\bibitem{DBLP:conf/icml/TrouillonWRGB16}
Trouillon, T., Welbl, J., Riedel, S., Gaussier, {\'{E}}., Bouchard, G.: Complex
  embeddings for simple link prediction. In: {ICML}. {JMLR} Workshop and
  Conference Proceedings, vol.~48, pp. 2071--2080. JMLR.org (2016)

\bibitem{tucker64extension}
Tucker, L.R.: {T}he extension of factor analysis to three-dimensional matrices.
  In: Gulliksen, H., Frederiksen, N. (eds.) {C}ontributions to mathematical
  psychology., pp. 110--127. Holt, Rinehart and Winston, New York (1964)

\bibitem{DBLP:conf/iclr/VashishthSNT20}
Vashishth, S., Sanyal, S., Nitin, V., Talukdar, P.P.: Composition-based
  multi-relational graph convolutional networks. In: {ICLR}. OpenReview.net
  (2020)

\bibitem{DBLP:conf/nips/VaswaniSPUJGKP17}
Vaswani, A., Shazeer, N., Parmar, N., Uszkoreit, J., Jones, L., Gomez, A.N.,
  Kaiser, L., Polosukhin, I.: Attention is all you need. In: {NIPS}. pp.
  5998--6008 (2017)

\bibitem{DBLP:conf/nips/VinyalsBLKW16}
Vinyals, O., Blundell, C., Lillicrap, T., Kavukcuoglu, K., Wierstra, D.:
  Matching networks for one shot learning. In: {NIPS}. pp. 3630--3638 (2016)

\bibitem{DBLP:conf/aaai/WangHLP19}
Wang, P., Han, J., Li, C., Pan, R.: Logic attention based neighborhood
  aggregation for inductive knowledge graph embedding. In: {AAAI}. pp.
  7152--7159. {AAAI} Press (2019)

\bibitem{DBLP:conf/nips/0004LSLLYA22}
Wang, R., Li, Z., Sun, D., Liu, S., Li, J., Yin, B., Abdelzaher, T.F.: Learning
  to sample and aggregate: Few-shot reasoning over temporal knowledge graphs.
  In: NeurIPS (2022)

\bibitem{DBLP:conf/emnlp/WuCCH20}
Wu, J., Cao, M., Cheung, J.C.K., Hamilton, W.L.: Temp: Temporal message passing
  for temporal knowledge graph completion. In: {EMNLP} {(1)}. pp. 5730--5746.
  Association for Computational Linguistics (2020)

\bibitem{DBLP:conf/emnlp/XiongYCGW18}
Xiong, W., Yu, M., Chang, S., Guo, X., Wang, W.Y.: One-shot relational learning
  for knowledge graphs. In: {EMNLP}. pp. 1980--1990. Association for
  Computational Linguistics (2018)

\bibitem{DBLP:conf/naacl/XuCNL21}
Xu, C., Chen, Y., Nayyeri, M., Lehmann, J.: Temporal knowledge graph completion
  using a linear temporal regularizer and multivector embeddings. In:
  {NAACL-HLT}. pp. 2569--2578. Association for Computational Linguistics (2021)

\bibitem{DBLP:conf/coling/XuNAYL20}
Xu, C., Nayyeri, M., Alkhoury, F., Yazdi, H.S., Lehmann, J.: Tero: {A}
  time-aware knowledge graph embedding via temporal rotation. In: {COLING}. pp.
  1583--1593. International Committee on Computational Linguistics (2020)

\bibitem{DBLP:journals/corr/YangYHGD14a}
Yang, B., Yih, W., He, X., Gao, J., Deng, L.: Embedding entities and relations
  for learning and inference in knowledge bases. In: {ICLR} (Poster) (2015)

\bibitem{DBLP:conf/cikm/ZhangZ0ZX022}
Zhang, F., Zhang, Z., Ao, X., Zhuang, F., Xu, Y., He, Q.: Along the time:
  Timeline-traced embedding for temporal knowledge graph completion. In:
  {CIKM}. pp. 2529--2538. {ACM} (2022)

\bibitem{DBLP:conf/aaai/ZhangDKSS18}
Zhang, Y., Dai, H., Kozareva, Z., Smola, A.J., Song, L.: Variational reasoning
  for question answering with knowledge graph. In: {AAAI}. pp. 6069--6076.
  {AAAI} Press (2018)

\end{thebibliography}

\appendix
\section{Implementation Details}
All experiments are implemented with PyTorch \cite{DBLP:conf/nips/PaszkeGMLBCKLGA19} on a single NVIDIA A40 with 48GB memory. We search hyperparameters following Table \ref{tab: hyperparameter search}. For each dataset, we do 108 trials to try different hyperparameter settings. We run 1000 episodes for each trail and compare their meta-validation results. We choose the setting leading to the best meta-validation result and take it as the best hyperparameter setting. The best hyperparameter setting is reported in Table \ref{tab: best param}. Our time-aware Transformer uses two heads and two attention layers for all experiments. The results of FITCARL is the average of five runs. The GPU memory usage, training time and the number of parameters are presented in Table \ref{tab: memory}, \ref{tab: train time} and \ref{tab: num of param}, respectively. For all datasets, we use all meta-training entities $\mathcal{E}'_{\text{meta-train}}$ as the considered unseen entities in each meta-training task $T$. This also applies during meta-validation and meta-test, where all the entities in $\mathcal{E}'_{\text{meta-valid}}$/$\mathcal{E}'_{\text{meta-test}}$ are considered appearing simultaneously in one evaluation task. All the datasets are taken from FILT's official repository\footnote{https://github.com/Jasper-Wu/FILT}.
We also take the pre-trained representations from it for our experiments. During evaluation, we follow previous RL-based TKG reasoning models TITer and CluSTeR and use beam search for answer searching. The beam size is 100 for all experiments.

We implement TITer with its official code\footnote{https://github.com/JHL-HUST/TITer}. We give it the whole background graph $\mathcal{G}_{\text{back}}$ as well as all meta-training quadruples $\mathbb{T}_{\text{meta-train}}$ for training. During meta-validation and meta-test, it is further given support quadruples for predicting the query quadruples.
\begin{table}[htbp]
    % \caption{Global caption}
    % \begin{minipage}{.435\linewidth}
    \caption{Hyperparameter searching strategy.}
        \label{tab: hyperparameter search}
      \centering
      % \resizebox{\linewidth}{!}{
        \begin{tabular}{ll}
            \toprule % <-- Toprule here
       \textbf{Hyperparameter} & \textbf{Search Space} \\
       \midrule % <-- Midrule here
       Embedding Size $d$& \{100, 200\}     \\
       Sampled Action Space Size &  \{25, 50, 100\}   \\
       Search Step $L$ &  \{3, 4\}\\
       Regularizer Coefficient $\eta$ & \{1e-11, 1e-9, 1e-7\}\\
       % $\lambda$ & \{0.2, 0.4\}\\
       Margin of Reward $\theta$ & \{1, 5, 10\}\\
    %   \# Negative Sample & \{8, 16, 32\}\\
       
      \bottomrule % <-- Bottomrule here
        \end{tabular}
        % }
    % \end{minipage}%
\end{table}
\begin{table}[htbp]
    \caption{Best hyperparameter settings.}
    \label{tab: best param}
    % \begin{minipage}{.565\linewidth}
      \centering
        % \resizebox{\linewidth}{!}{
        \begin{tabular}{lccc} 
      \toprule % <-- Toprule here
     \multicolumn{1}{l}{\textbf{Datasets}} & \multicolumn{1}{c}{\textbf{ICEWS14-OOG}} & \multicolumn{1}{c}{\textbf{ICEWS18-OOG}} &\multicolumn{1}{c}{\textbf{ICEWS0515-OOG}} \\
      \midrule % <-- Midrule here
      \textbf{Hyperparameter} &   &  &   \\
       \midrule % <-- Midrule here
       Embedding Size $d$ & 100     &  100&  100 \\
       Sampled Action Space Size &  50   &  50&   50 \\
       Search Step $L$ &  3 &  3 & 3\\
       Regularizer Coefficient $\eta$ & 1e-9& 1e-9 & 1e-9\\
       Margin of Reward $\theta$ & 5 & 5 & 5\\
    %   \# Negative Sample & 32 & 32 & 32 \\
       
      \bottomrule % <-- Bottomrule here
    \end{tabular} 
    % }

    % \end{minipage} 
\end{table}

\begin{table}[htbp]
\caption{GPU memory usage (MB).}
    \label{tab: memory}
    % \caption{Global caption}
    % \begin{minipage}{.5\linewidth}
      \centering
      % \resizebox{\linewidth}{!}{
      \begin{tabular}{lcccccc} 
      \toprule % <-- Toprule here
     \multicolumn{1}{l}{\textbf{Datasets}} & \multicolumn{2}{c}{\textbf{ICEWS14-OOG}} & \multicolumn{2}{c}{\textbf{ICEWS18-OOG}} &\multicolumn{2}{c}{\textbf{ICEWS0515-OOG}} \\
      \cmidrule(lr){2-3}\cmidrule(lr){4-5}\cmidrule(lr){6-7}
       & \multicolumn{2}{c}{GPU Memory}  & \multicolumn{2}{c}{GPU Memory} & \multicolumn{2}{c}{GPU Memory}  \\
      \textbf{Model} & 1-S & 3-S & 1-S & 3-S & 1-S & 3-S\\
       \midrule
       FITCARL & 10729 & 11153 & 14761 & 15419 & 14765 & 15475 \\
       
      \bottomrule % <-- Bottomrule here
    \end{tabular}
        % }
        
    % \end{minipage}%
\end{table}
\begin{table}[htbp]
\caption{Training time (min).}
    \label{tab: train time}
    % \caption{Global caption}
    % \begin{minipage}{.5\linewidth}
      \centering
      % \resizebox{\linewidth}{!}{
      \begin{tabular}{lcccccc} 
      \toprule % <-- Toprule here
     \multicolumn{1}{l}{\textbf{Datasets}} & \multicolumn{2}{c}{\textbf{ICEWS14-OOG}} & \multicolumn{2}{c}{\textbf{ICEWS18-OOG}} &\multicolumn{2}{c}{\textbf{ICEWS0515-OOG}} \\
      \cmidrule(lr){2-3}\cmidrule(lr){4-5}\cmidrule(lr){6-7}
       & \multicolumn{2}{c}{Time}  & \multicolumn{2}{c}{Time} & \multicolumn{2}{c}{Time}  \\
      \textbf{Model} & 1-S & 3-S & 1-S & 3-S & 1-S & 3-S\\
       \midrule
       FITCARL & 225 & 85 & 305 & 764 & 1059 & 297 \\
       
      \bottomrule % <-- Bottomrule here
    \end{tabular}
        % }
        
    % \end{minipage}%
\end{table}
\begin{table}[htbp]
\caption{Number of parameters.}
    \label{tab: num of param}
    % \begin{minipage}{.5\linewidth}
      \centering
        % \resizebox{\linewidth}{!}{
        \begin{tabular}{lcccccc} 
      \toprule % <-- Toprule here
     \multicolumn{1}{l}{\textbf{Datasets}} & \multicolumn{2}{c}{\textbf{ICEWS14-OOG}} & \multicolumn{2}{c}{\textbf{ICEWS18-OOG}} &\multicolumn{2}{c}{\textbf{ICEWS0515-OOG}} \\
      \cmidrule(lr){2-3}\cmidrule(lr){4-5}\cmidrule(lr){6-7}
       & \multicolumn{2}{c}{\# Param}  & \multicolumn{2}{c}{\# Param} & \multicolumn{2}{c}{\# Param} \\
      \textbf{Model} & 1-S & 3-S & 1-S & 3-S & 1-S & 3-S\\
       \midrule
       FITCARL & 8271206 & 8271410 & 14633206 & 10006710 & 9615206 & 9615410 \\
       
      \bottomrule % <-- Bottomrule here
    \end{tabular} 
        
    % }
    
    % \end{minipage} 
\end{table}
\section{Difference between TKGC and TKG forecasting}
Assume we have a TKG $\mathcal{G} = \{(s,r,o,t)|s,o \in \mathcal{E}, r \in \mathcal{R}, t \in \mathcal{T}\} \subseteq \mathcal{E} \times \mathcal{R} \times \mathcal{E} \times \mathcal{T}$, where $\mathcal{E}$, $\mathcal{R}$, $\mathcal{T}$ denote a finite set of entities, relations and timestamps, respectively. We define the TKG forecasting task (also known as TKG extrapolation) as follows. Assume we have an LP query $(s_q,r_q,?,t_q)$ (or $(?,r_q,o_q,t_q)$) derived from a query quadruple $(s_q,r_q,o_q,t_q)$.
% , and we denote all the ground-truth quadruples as $\mathcal{F}$. 
TKG forecasting aims to predict the missing entity in the LP query, given the observed \textbf{past} TKG facts $\mathcal{O} = \{(s_i, r_i, o_i, t_i)|t_i < t_q\}$. Such temporal restriction is not imposed in TKGC (also known as TKG interpolation), where the observed TKG facts from any timestamp, including $t_q$ and the timestamps after $t_q$, can be used for prediction.

TITer is designed for TKG forecasting, therefore it only performs its RL search process in the direction pointing at the past. This leads to a great loss of information along the whole time axis. TITer also does not use a meta-learning framework for adapting to the few-shot setting, which is also a reason for its weak performance on TKG few-shot OOG LP. Please refer to the papers studying TKG forecasting, e.g., \cite{DBLP:journals/corr/abs-1904-05530,DBLP:conf/emnlp/HanDMGT21}, for more details.

\section{Ablation E Details}
We describe here how we change equations in ablation E to build a model variant without using any temporal information. First, we change the action space sampling method to random sample, which corresponds to ablation A1. This means we do not use temporal information to compute time-adaptive sampling probabilities. Next, we neglect the last term in Equation 2 of the main paper. It thus becomes
\begin{equation}
\label{eq: time att transformer static}
\begin{aligned}
    &\text{att}_{u,v} = \frac{\text{exp}(\alpha_{u,v})}{\sum_{k=1}^{K+1} \text{exp}(\alpha_{u,k})},\\
    &\alpha_{u,v} = \frac{1}{\sqrt{d}}( \mathbf{W}_{TrQ} \mathbf{h}_u)^\top(\mathbf{W}_{TrK} \mathbf{h}_v),
\end{aligned}    
\end{equation}
which corresponds to ablation D. Finally, we remove every term in all equations containing time-difference representations. For a node $(e,t)$, its representation becomes $\mathbf{h}_e$. Thus, Equation 3 of the main paper becomes
\begin{equation}
\begin{aligned}
        &\mathbf{h}_{\text{hist}^{(l)}} = \text{GRU}\left( \left(\mathbf{h}_{{r}^{(l)}} \| \mathbf{h}_{e^{(l)}}\right),\mathbf{h}_{\text{hist}^{(l-1)}}\right),\\
        &\mathbf{h}_{\text{hist}^{(0)}} = \text{GRU}\left( \left(\mathbf{h}_{{r}_{\text{dummy}}} \| \mathbf{h}_{e'}\right), \mathbf{0}\right).
\end{aligned}
\end{equation}
Equation 4 of the main paper becomes
\begin{equation}
\label{eq: attentional feature static}
\begin{aligned}
        &\mathbf{h}_{\text{hist}^{(l)}, q|a} = \text{att}_{\text{hist}^{(l)}, a} \cdot \Bar{\mathbf{h}}_{\text{hist}^{(l)}} + \text{att}_{q, a} \cdot \Bar{\mathbf{h}}_{q}, \\
        &\Bar{\mathbf{h}}_{\text{hist}^{(l)}} = {\mathbf{W}_1}^\top\mathbf{h}_{\text{hist}^{(l)}}, \quad
        \Bar{\mathbf{h}}_{q} = {\mathbf{W}_2}^\top\left(\mathbf{h}_{r_q}\|\mathbf{h}_{e'}\right).
\end{aligned}
\end{equation}
Equation 5 and 6 of the main paper become
\begin{equation}
    \text{att}_{\text{hist}^{(l)},a} = \frac{\text{exp}(\phi_{\text{hist}^{(l)},a})} {\text{exp}(\phi_{\text{hist}^{(l)},a})+\text{exp}(\phi_{q,a})}, \text{att}_{q,a} = \frac{\text{exp}(\phi_{q,a})}{\text{exp}(\phi_{\text{hist}^{(l)},a})+\text{exp}(\phi_{q,a})},
\end{equation}
where
\begin{equation}
    \begin{aligned}
        \phi_{\text{hist}^{(l)},a} = {\Bar{\mathbf{h}}_a}^\top \Bar{\mathbf{h}}_{\text{hist}^{(l)}}, \ 
        \phi_{q,a} = {\Bar{\mathbf{h}}_a}^\top \Bar{\mathbf{h}}_{q},\\
        \Bar{\mathbf{h}}_a = {\mathbf{W}_3}^\top\left(\mathbf{h}_{r_a}\|\mathbf{h}_{e_a}\right).
    \end{aligned}
\end{equation}
Equation 8 of the main paper becomes
\begin{equation}
\label{eq: confidence static}
\begin{aligned}
        &\text{conf}_{a|q} = \frac{\text{exp}(\psi_{a|q})}{\sum_{a' \in \mathcal{A}^{(l)}} \text{exp}(\psi_{a'|q})}, \ \text{where}\ \psi_{a|q} = \mathcal{W} \times_1 \mathbf{h}_{e'} \times_2 \mathbf{h}_{r_q} \times_3 \mathbf{h}_{e_a}.
\end{aligned}
\end{equation}
The other equations remain unchanged. To this end, we create a model variant that uses no temporal information.

\section{Evaluation Metrics}
We use two evaluation metrics, i.e., mean reciprocal rank (MRR) and Hits@1/3/10. For every LP query $q$, we compute the rank $rank_q$ of the ground truth missing entity. We define MRR as: $\frac{1}{\sum_{e' \in \mathcal{E}'_{\text{meta-test}}} |Que_{e'}|}\sum_{e' \in \mathcal{E}'_{\text{meta-test}}} \sum_{q \in Que_{e'}} \frac{1}{rank_q}$. Hits@1/3/10 denote the proportions of the predicted links where ground truth missing entities are ranked as top 1, top3, top10, respectively. We also use the filtered setting proposed in \cite{DBLP:conf/nips/BordesUGWY13} for fairer evaluation.

\section{Meta-Training Algorithm of FITCARL}
\begin{algorithm}[t]
% \tiny
% \small
\scriptsize
% \small
\caption{FITCARL Meta-Training}
\label{alg: one shot}

\DontPrintSemicolon
\KwInput{Meta-training entities $\mathcal{E}'_{\text{meta-train}}$, background TKG $\mathcal{G}_\text{back}$, shot size $K$}
% \KwInput{Background graph $\mathcal{G}'$}
% \KwInput{Model parameters $\theta$}

\For{\textup{episode =} 1: M}
{
\For{$e' \in \mathcal{E}'_{\text{meta-train}}$}
{
% Sample a support quadruple to make the support set $\mathcal{S}_r = (s_0,r,o_0,t_0)$
Sample a support set $Sup_{e'}$ and a query set $Que_{e'}$

% \If {\textup{One-Shot Interpolated LP}}
% {Sample a batch of query quadruples $\mathcal{Q}_r = \{(s_q,r,o_q,t_q)\}$}

% \Else(\tcp*[h]{One-Shot Extrapolated LP})
% {Sample a batch of query quadruples $\mathcal{Q}_r = \{(s_q,r,o_q,t_q) | t_0 < t_q\}$}
Learn meta-representations $\{\mathbf{h}_{e'}^i\}_{i=1}^K$
}
% Derive LP query $q$ from each query quadruple in $Que_{e'}$ and compute $\mathbf{h}_{e'|q}$ \tcp*[h]{Section \ref{sec: unseen entity learning}}
\For{$e' \in \mathcal{E}'_{\text{meta-train}}$}
{
\For{query  $\in Que_{e'}$}
{
Derive LP query $q$ from $query$

Compute $\mathbf{h}_{e'}$ using time-aware Transformer \tcp*[h]{Section \ref{sec: unseen entity learning}}

Initialize $s^{(0)} \leftarrow (e', t_q, e', r_q, t_q)$

$\{R(s^{(l)}, a^{(l)})\}_{l=0}^{L-1}, \{\mathcal{L}_{\text{KL}|q}^{(l)}\}_{l=0}^{L-1}, \{\pi(a^{(l)}|s^{(l)})\}_{l=0}^{L-1}\leftarrow$ Search$(L,s^{(0)})$
}
}

Compute loss $\mathcal{L}_T$ \tcp*[h]{Equation \ref{eq: loss}}

Update model parameters using gradient of $\bigtriangledown	\mathcal{L}_T$
% Compute meta-information $\mathbf{h}^{\text{meta}}_{(s_0,r,o_0,t_0)}$

% Learn meta-representation $\mathbf{h}_{r}$ with meta-representation learner

% Pollute each $q \in \mathcal{Q}_r$ and generate polluted quadruples $\{q^-\}$ 

% Compute time-aware representations for entities in all $q$ and $\{q^-\}$ \tcp*[h]{Equation \ref{eq: queryemb}}

% Compute scores for all $q$ and $\{q^-\}$ with metric function $\psi$

% Calculate the loss $\mathcal{L}$

% Update model parameters using gradient of loss $\bigtriangledown	\mathcal{L}$
}

\Fn{\textup{Search}$(L,s^{(0)})$}
{
\For{l \textup{=} 0:L-1}
{
Sample action space $\mathcal{A}^{(l)}$ from all observed outgoing edges of node $(e^{(l)}, t^{(l)})$

Compute $P(a|s^{(l)}, \text{hist}^{(l)})$ and $\text{conf}_{a|q}$ for $a \in \mathcal{A}^{(l)}$ \tcp*[h]{Equation \ref{eq: prob before confidence}, \ref{eq: confidence}}

Compute $\pi(a|s^{(l)})$ for each $a \in \mathcal{A}^{(l)}$ \tcp*[h]{Equation \ref{eq: policy}}

Compute $\mathcal{L}_{\text{KL}|q}^{(l)}$ \tcp*[h]{Equation \ref{eq: kl}}

Sample $a^{(l)} = (e_{a^{(l)}}, r_{a^{(l)}}, t_{a^{(l)}})$ according to policy $\pi$

Compute reward $R(s^{(l)}, a^{(l)})$

Execute $a^{(l)}$, agent transfers to state $s^{(l+1)} = (e^{(l+1)}, r^{(l+1)}, e', r_q, t_q)$

}
return $\{R(s^{(l)}, a^{(l)})\}_{l=0}^{L-1}, \{\mathcal{L}_{\text{KL}|q}^{(l)}\}_{l=0}^{L-1}, \{\pi(a^{(l)}|s^{(l)})\}_{l=0}^{L-1}$\;
}
% \Fn{\textup{Sample Task}}
% {
% \For{$e' \in \mathcal{E}'_{\text{meta-train}}$}
% {
% % Sample a support quadruple to make the support set $\mathcal{S}_r = (s_0,r,o_0,t_0)$
% Sample a support set $Sup_{e'}$ and a query set $Que_{e'}$
% }
% return 0\;
% }
\end{algorithm}
We train FITCARL with episodic training. We present our meta-training process in Algorithm \ref{alg: one shot}.
\end{document}